\documentclass{article}

\PassOptionsToPackage{numbers, compress}{natbib}
\usepackage[main, final]{neurips_2026}

\usepackage[utf8]{inputenc} 
\usepackage[T1]{fontenc}    
\usepackage{hyperref}       
\usepackage{url}            
\usepackage{booktabs}       
\usepackage{amsfonts}       
\usepackage{nicefrac}       
\usepackage{microtype}      
\usepackage{xcolor}         
\usepackage{graphicx}
\usepackage{enumitem}
\usepackage{amsmath}
\usepackage{algorithm}
\usepackage{algpseudocode}
\usepackage{multirow}
\usepackage{makecell}
\usepackage{longtable}
\usepackage{float}
\usepackage{titlesec}
\usepackage{amssymb}
\usepackage{parskip}
\usepackage{fancyvrb}
\usepackage{fvextra}
\usepackage[most]{tcolorbox}

\titlespacing*{\paragraph}{0pt}{1ex plus 0.5ex minus 0.1ex}{0.5em}

\title{Agentic Neural Architecture Search}

\author{%
  Seokhoon Jeong$^{1}$ \quad Mijung Kim$^{2}$ \quad Taehwan Kim$^{1}$ \\\\
  $^{1}$Artificial Intelligence Graduate School \quad
  $^{2}$Department of Computer Science and Engineering \\
  Ulsan National Institute of Science and Technology \\
  \texttt{\{shjd0246, mijungk, taehwankim\}@unist.ac.kr}
}

\begin{document}

\maketitle

\begin{abstract}
Neural architecture search (NAS) methods have grown increasingly efficient, but they still operate within manually designed search spaces that require domain expertise and must be rebuilt for each new task. Large language models (LLMs) can write architectures in an open-ended code space, yet it is unclear how the work should be divided between LLM-driven design and NAS-driven search. We propose a way to combine the two. An LLM first designs a seed architecture and then decomposes it into a \emph{slotted architecture}: a scaffold with named, interchangeable module slots that defines a bounded, task-specific search space, which a conventional NAS algorithm then explores without any manual engineering. We implement this in \textbf{AgentNAS}, a three-phase pipeline whose phases can be evaluated separately. On 17 tasks from NAS-Bench-360 and Unseen NAS, covering classification, dense regression, segmentation, and multi-label tagging, AgentNAS outperforms published baselines, including task-specific expert designs, on 11 tasks. Extensive ablation studies show that LLMs find strong architectures quickly but soon plateau, and that switching to NAS at this point yields further gains, indicating that the two approaches are complementary. Experiments with five LLMs show that seed architecture design is effective across all of them, whereas constructing a usable search space requires a minimum level of model capability. Our code is available at \url{https://github.com/alroimfebruary/AgentNAS}.
\end{abstract}
\section{Introduction}
\label{sec:intro}
Neural architecture search (NAS) aims to replace the manual design of neural networks with an automated search. Weight-sharing methods~\citep{liu2018darts,pham2018efficient} and evolutionary algorithms~\citep{real2019regularized} have made the search itself increasingly efficient. Every such method, however, still presupposes a hand-engineered search space that demands domain expertise and must be rebuilt for each new task~\citep{elsken2019neural,white2023neural}. Large language models (LLMs) have begun to loosen this constraint. Having absorbed architectural knowledge from the research literature, they can generate or mutate network code in an open-ended space, and have been used as mutation operators in evolutionary search~\citep{chen2023evoprompting,morris2024llm,nasir2024llmatic,zhu2025llm}, as optimizers~\citep{zheng2023can}, and as hyperparameter tuners~\citep{zhang2023using,liu2024large}.
In these methods the LLM both \emph{expands} the space of possible architectures and \emph{searches} it, which makes it hard to tell when each of the two capabilities adds value.
 
This leaves two questions open: can an LLM replace NAS, and if not, how should the work be divided between them? An LLM can propose diverse architectures from its prior, but each proposal is close to an independent sample and does not systematically test how several design choices interact. NAS is built for exactly this kind of combinatorial search, but it needs a bounded space to search in. Studying how the two interact therefore requires a framework that (i)~lets an LLM construct the search space, (ii)~lets a conventional NAS algorithm explore it, and (iii)~keeps the two apart so that the contribution of each can be measured.
 
We propose such a framework. An LLM first designs a seed architecture and then decomposes it into a \emph{slotted architecture}, a scaffold with named, interchangeable module slots that defines a bounded, task-specific search space for conventional NAS. The decomposition keeps the macro-level decisions of the LLM, such as depth, width progression, and backbone type, and leaves the module-level choices to NAS: each slot holds several alternative modules, and NAS searches for the combination across slots that works best together. We implement this framework in \textbf{AgentNAS}, a three-phase pipeline in which the seed search, the search space construction, and the NAS step can each be evaluated on its own.
 
We evaluate AgentNAS on 17 tasks from NAS-Bench-360~\citep{tu2022bench} and Unseen NAS~\citep{geada2024insights}, covering classification, dense regression, segmentation, and multi-label tagging. AgentNAS outperforms published baselines, including task-specific expert designs, on 11 tasks. Ablations on six representative tasks show how the two components divide the work. LLM sampling improves the architecture quickly but saturates, and running NAS from the saturation point improves further. Separate controls show that these gains come mainly from the architecture rather than from the training recipe the LLM selects. Across five different LLMs, seed design works with every model, while weaker models sometimes fail to construct a usable search space.  
 
Our contributions are as follows:
\begin{enumerate}[leftmargin=*]
    \item We present \textbf{AgentNAS}, a framework in which an LLM decomposes its own generated architecture into a task-specific search space that standard NAS algorithms can explore without manual engineering.
    \item We study the division of labor between LLM-driven design and NAS. We define complementarity in terms of budget allocation and separate the effect of the architecture from that of the training recipe.
    \item AgentNAS outperforms published baselines on 11 of 17 diverse tasks, including the blind Unseen NAS setting, in which the LLM receives no information about the task domain.
\end{enumerate}
\section{Related Work}
\label{sec:related}
\subsection{Neural Architecture Search}

Neural architecture search (NAS) automates the design of neural network architectures, a process that has traditionally relied on human trial and error. The earliest neuroevolution methods evolved the weights and topology of neural networks with genetic algorithms~\citep{MillerTH89,Yao99,Stanley2002EvolvingNN,WhitesonS06}. In the deep learning era, \citet{zoph2016neural} first showed that a recurrent controller trained with reinforcement learning can find architectures competitive with hand-designed models, although at a very large computational cost. Two main lines of research followed. Weight-sharing methods~\citep{bender2018understanding,bender2020can,cai2018proxylessnas,liu2018darts,pham2018efficient,xu2019pc} train a single over-parameterized supernet once and evaluate candidate architectures as its sub-networks. Evolutionary methods~\citep{cai2019once,lu2019nsga,real2017large,real2019regularized} maintain a population of candidates and improve it through mutation and selection. Both lines, however, search within a fixed search space designed by humans.

Tabular benchmarks with pre-computed architecture performance~\citep{ying2019bench,dong2020bench,dong2021nats} reinforced this tendency. Because they are built on cell-based convolutional search spaces for image classification~\citep{krizhevsky2009learning,deng2009imagenet}, most later NAS research has been evaluated in the same narrow setting. Other work has applied NAS beyond it, to transformers~\citep{so2019evolved,yang2023evolutionary,li2021bossnas}, graph neural networks~\citep{gao2019graphnas,li2020autograph,zhou2022auto}, and generative adversarial networks~\citep{gong2019autogan,gao2020adversarialnas}.

Whatever the search strategy or target domain, these methods share one limitation: the search space must be engineered by hand. Several studies have found that the expressiveness of the search space matters at least as much as the choice of search algorithm~\citep{geada2020bonsai,liu2019auto,liu2017hierarchical,roberts2021rethinking,ru2020neural}, yet designing such spaces still takes substantial effort and domain expertise. \citet{real2020automl} avoided a predefined space by composing models from basic mathematical operations, at a prohibitive computational cost. Search spaces defined by context-free grammars~\citep{schrodi2023construction,ericsson2024einspace} are far more expressive, but they are difficult to combine with supernet-based search, and they move the manual effort from designing architectures to designing grammars. In all of these approaches, the search space remains a manual design artifact.

\subsection{Large Language Models for Neural Architecture Search}

Large language models (LLMs) offer a way around this limitation. Instead of searching within a predefined space, an LLM can write or modify architecture code directly, so the space of reachable architectures is essentially unbounded. Most often, LLMs act as code-level mutation and crossover operators within evolutionary search~\citep{chen2023evoprompting,morris2024llm,nasir2024llmatic,zhu2025llm}, or as architecture generators that produce candidate networks from natural-language or code specifications~\citep{yang2025nader,rahman2024lemo}. They have also been used as optimizers that propose the next candidate from the search history~\citep{zheng2023can,yu2023gpt}, as rerankers~\citep{hu2025lm}, as extractors of design principles~\citep{zhou2025design}, and as hyperparameter tuners~\citep{zhang2023using,liu2024large}. A related line of work trains a generative model on architecture data instead of using a general-purpose LLM. In all of these works, the LLM operates inside the search loop as an operator or optimizer. To our knowledge, no prior work uses an LLM to construct a bounded, task-specific search space and then hands that space to a conventional NAS algorithm. As a result, how the work should be divided between LLM-driven design and NAS-driven search, and when each part adds value, has not been studied empirically.

\subsection{Automated Scientific Discovery}

More broadly, combining LLMs, multi-agent systems, programmatic evaluators, and evolutionary search has been used to discover new algorithms~\citep{romera2024mathematical,novikov2025alphaevolve,press2026algotune} and to carry out scientific research autonomously~\citep{lu2024ai,yamada2025ai,schmidgall2025agentrxiv,schmidgall2025agent}. This approach has also been applied to scientific machine learning, for example to find PINN configurations and PDE surrogates~\citep{wuwu2025pinnsagent,song2026can,jiang2026agenticsciml}, while concurrent work enlarges the design space of neural operators with architectures based on state-space models~\citep{tiwari2025latent,song2026adaptive,song2025adaptivefourier}. These efforts automate architecture design for scientific computing. AgentNAS addresses general-purpose NAS and applies to such problems as well, as our Darcy Flow experiments show.
\section{Method}
\label{sec:method}

\begin{figure*}[t!]
  \centering
  \includegraphics[width=\linewidth]{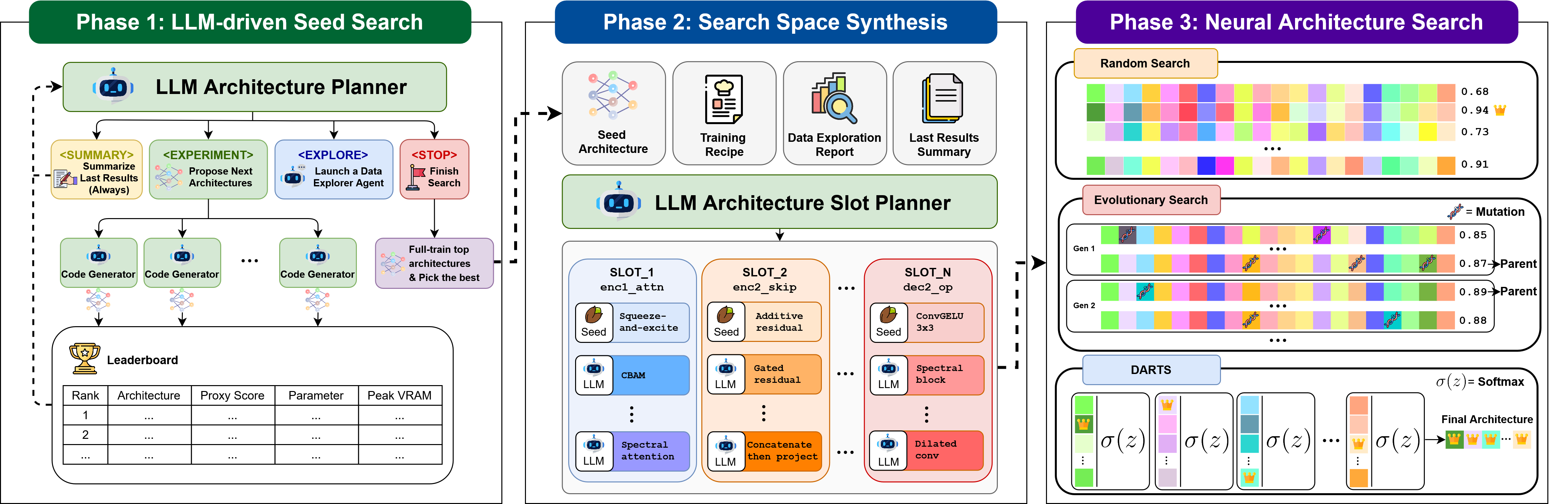}
  \caption{The AgentNAS pipeline. Phase~1: an LLM Planner repeatedly proposes architectures, which are implemented, trained under a proxy schedule, and added to a leaderboard. The Planner summarizes each round of results and continues until it stops improving, returning a seed architecture and its training recipe. Phase~2: the seed is decomposed into a slotted architecture with several alternatives per slot, which defines a task-specific search space. Phase~3: a conventional NAS algorithm explores this space. Each square represents one slot, and a full row of squares is one candidate architecture. The panel shows random search, regularized evolution, and differentiable search as examples.}
  \label{fig:pipeline}
\end{figure*}

\begin{figure*}[t!]
  \centering
  \includegraphics[width=\linewidth]{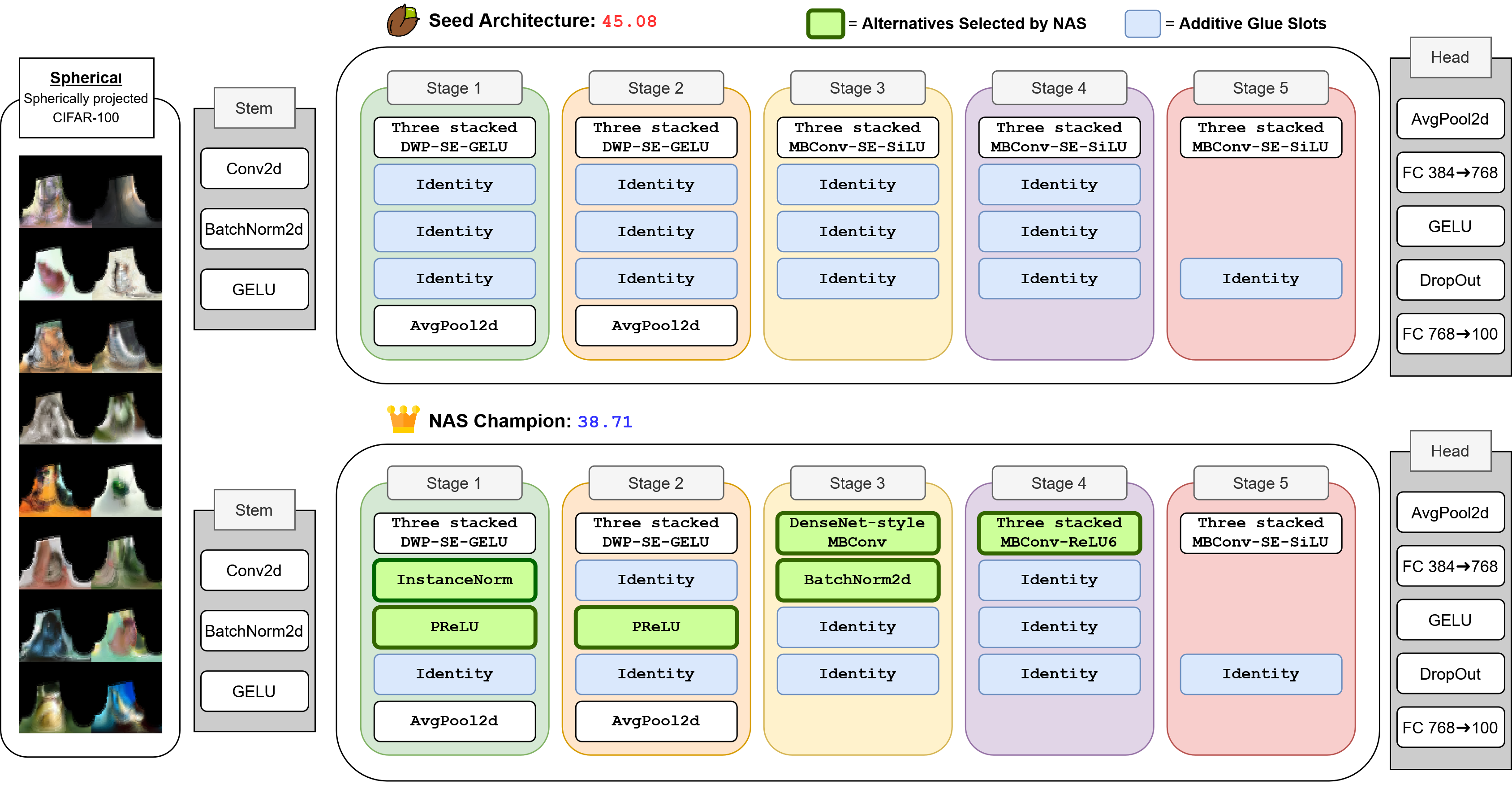}
  \caption{Slotted architecture for the \textbf{Spherical} task. Module-level slots, such as \emph{three stacked DWP-SE-GELU} blocks, can be replaced as a whole, while additive glue slots between modules default to identity and can add a normalization or activation. Starting from the seed (top, 45.08\% test error), NAS replaces the modules of Stages~3 and~4 and adds InstanceNorm and PReLU in the early stages, reaching 38.71\% (bottom).}
  \label{fig:slotted}
\end{figure*}

AgentNAS is a three-phase pipeline in which an LLM designs a seed architecture and the search space around it, and a conventional NAS algorithm then searches that space. \citet{ericsson2024einspace} found that starting NAS from human-designed architectures helps; we replace the human designer with an LLM and let the LLM also define the search space. Figure~\ref{fig:pipeline} gives an overview, and Figure~\ref{fig:slotted} shows the slotted architecture for the Spherical task. The prompts of all agents are listed in Appendix~\ref{app:context} and further implementation details in Appendix~\ref{app:impl}.

\subsection{Phase~1: LLM-Driven Seed Search}
\label{sec:phase1}

Phase~1 is a loop between three agents. At each step, a \emph{Planner} chooses one of three actions: \texttt{<EXPLORE>} launches a \emph{Data Explorer} that analyzes the dataset by running code in a sandbox, \texttt{<EXPERIMENT>} proposes up to eight architectures together with their training recipes (optimizer, learning rate, schedule, augmentation, batch size, and epochs), and \texttt{<STOP>} ends the search. A \emph{Code Generator} implements each proposed architecture in PyTorch, and the candidate is trained under a proxy schedule of $30\%$ of the proposed epochs, with a one-hour GPU-time cap and the 11\,GB VRAM limit of our GPUs. After each round, the Planner receives the updated leaderboard and writes a summary of what worked and what failed, which it uses to plan the next round. Phase~1 ends when the Planner stops, after 160 evaluations, or after 64 successful evaluations without improvement. Other resource constraints, such as a parameter ceiling, can be imposed in the same way, through the prompt and by rejecting candidates that violate them (Appendix~\ref{app:param_constrained}).

After the search, the eight best candidates by proxy validation metric are trained with the full schedule, and the one with the best validation metric becomes the \emph{seed architecture}. We report it as AgentNAS-LLM. Selection uses only validation metrics; test metrics are never shown to the pipeline.

\subsection{Phase~2: Slotted Architecture Synthesis}
\label{sec:phase2}

Phase~2 turns the seed into a \emph{slotted architecture}: a scaffold of the seed in which selected components are replaced by named slots, each with several interchangeable alternatives. The scaffold keeps most of the macro-level decisions of the seed, such as depth, width progression, stage structure, and backbone type, and exposes choices at the module level for NAS. A \emph{Slotted Architecture Planner} receives the source code and validation performance of the seed, the Data Explorer's report, and the Planner's summaries, and produces the scaffold together with up to 20 slots of up to eight alternatives each. The Code Generator implements every alternative as a self-contained \texttt{nn.Module}, and each one must pass a validation check before the search starts, with up to ten repair attempts.

Creating a slot for every operation would make the search space far too large (up to ${\sim}10^{90}$ configurations in a preliminary experiment). Following hierarchical~\citep{tan2019mnasnet,wu2019fbnet} and cell-based NAS~\citep{liu2018darts}, which obtain good architectures by stacking well-designed blocks, we instead use the slot types described below: the LLM designs coherent block-level alternatives, and NAS searches over their combinations.

\paragraph{Modularization.}
Some seeds are written as a flat stack of primitives, with convolution, normalization, and activation applied one after another. In that case, a \emph{Modularizer} first groups the primitives of each stage into a self-contained module, and the refactored seed is checked to produce the same outputs as the original. Without this step, the Slotted Architecture Planner sometimes cuts the seed into very small pieces; this allows topological changes but makes the search unstable (Appendix~\ref{app:modularization}).

\paragraph{Module-level slots.}
A module-level slot covers one coherent unit of the network, typically a stage. Each alternative is a complete module that performs the same role with a different design, so this slot type changes the seed by \emph{replacing} whole components while keeping each block internally consistent.

\paragraph{Additive glue slots.}
Architectures found by NAS often contain components that a human designer would not add. To allow such additions, the Slotted Architecture Planner places \emph{additive} glue slots between module-level slots, applied to the output of the preceding module. Each glue slot defaults to identity, so the seed is unchanged unless NAS selects an alternative. Glue slots let NAS insert extra operations, normalizations, activations, or skip connections without changing the internals of neighboring modules.

\paragraph{Learning rate slot.}
The training recipe from Phase~1 is tuned for the seed, but replacing modules can change the model size considerably, and with it the best learning rate. We therefore add a virtual slot that scales the seed's learning rate by $\mu\!\in\!\{0.25,0.5,1.0,2.0,4.0\}$. All other recipe settings stay fixed. NAS selects a different learning rate on several tasks (Appendix~\ref{app:recipe}).

\subsection{Phase~3: Slotted Neural Architecture Search}
\label{sec:phase3}

A slotted architecture is a discrete, categorical search space, so most NAS algorithms can search it directly. We use regularized evolution~\citep{real2019regularized} by default, with a population of 16 initialized with the seed, its single- and double-slot mutations, and random configurations. Phase~3 has its own budget of 160 evaluations and the same patience rule as Phase~1. Appendix~\ref{app:search_algorithm} compares regularized evolution with random search~\citep{li2020random} and GDAS~\citep{dong2019searching}, a Gumbel-Softmax variant of DARTS~\citep{liu2018darts}. Candidates are trained with the proxy schedule of Phase~1, and the eight best candidates are then fully trained and selected by validation metric in the same way. Implementation details are in Appendix~\ref{app:impl-search}.
\section{Experiments}
\label{sec:experiments}

\subsection{Setup}
\label{sec:setup}

\paragraph{Benchmarks.}
We evaluate on 17 tasks from two benchmark suites. NAS-Bench-360~\citep{tu2022bench} contributes 10 tasks covering 1D and 2D classification, multi-label audio tagging (FSD50K), dense regression (Darcy Flow, PSICOV), and segmentation (Cosmic). Unseen NAS~\citep{geada2024insights} contributes 7 classification tasks built to test domain-agnostic architecture search, such as predicting chess game outcomes from board states (Chesseract) and attributing authorship from character-level encodings (Gutenberg). Unseen NAS was run as a competition in which participants could not see the evaluation datasets, and data shapes and split sizes were chosen to resemble well-known benchmarks so that the domain could not be guessed. We follow this blind setting: the LLM receives only input and output shapes, the loss, and the metric direction, under the placeholder name \texttt{TASK\_X}. It can still inspect the data through the Data Explorer, but we observe that it identifies the true domain only for Chesseract (Table~\ref{tab:blind_recognition}). Unseen NAS further distinguishes Type-1 tasks, for which an expert could design a strong architecture from domain knowledge, from Type-2 tasks, for which this is nearly impossible without task-specific tools; Language, Gutenberg, and GeoClassing are Type-2. We exclude the Type-2 task Isabella because the rights holder does not permit redistribution. Details are in Appendix~\ref{app:benchmarks}.

\paragraph{Baselines.}
On NAS-Bench-360, we compare with WideResNet-16-4~\citep{zagoruyko2016wide}, GAEA DARTS, the expert-designed architecture provided by the benchmark~\citep{tu2022bench}, and einspace~\citep{ericsson2024einspace} with regularized evolution from WideResNet-16-4, the previous state of the art on this suite. On Unseen NAS, we compare with ResNet-18~\citep{he2016deep}, the per-task best of six conventional NAS methods reported by \citet{geada2024insights} (PC-DARTS~\citep{xu2019pc}, DrNAS~\citep{chen2020drnas}, Bonsai-Net~\citep{geada2020bonsai}, random search on the DARTS and Bonsai-Net spaces, and regularized evolution on hNB201), and the per-task best of all einspace variants. All baseline numbers are taken from the original papers; the only number we reproduce is the fixed-seed column of Table~\ref{tab:agentnas_fixedseed}. Per-method results are in Appendix~\ref{app:baselines}.

\paragraph{Search budget.}
AgentNAS evaluates at most 160 architectures in Phase~1 and 160 in Phase~3, 320 in total; either phase can stop earlier through the patience rule or the Planner's \texttt{<STOP>} action. einspace evaluates 500 architectures when it starts from a seed (ResNet-18, WRN16-4, or a mix of seeds) and 1{,}000 when it searches from scratch or by random search. Supernet-based methods (GAEA DARTS, PC-DARTS, DrNAS) train a single shared network, so their cost cannot be expressed as a number of evaluations; we compare GPU hours and wall-clock time in Appendix~\ref{app:efficiency}. The per-task best baselines on Unseen NAS take the maximum over methods with different budgets.

\paragraph{LLMs and compute.}
All agents use Claude Sonnet~4.6~\citep{anthropic2026claude} at temperature $0.7$, each with its own context window, and the same prompts are used for all tasks without per-task tuning. The ablations additionally use Claude Haiku~4.5 and Claude Opus~4.6, and Appendix~\ref{app:llm_families} adds GPT-5.4 and the open-weight Qwen3.6-35B-A3B. Each architecture is trained on a single RTX~2080\,Ti with 11\,GB of VRAM, on a node of eight GPUs.

\paragraph{Reporting.}
Tables~\ref{tab:nas360_agentnas} and~\ref{tab:unseen_agentnas} report a single run per task with random seed 42; the ablation tasks are additionally run with seeds 43--51 (Table~\ref{tab:multiseed}). All selections are based on validation metrics, and we report the test metric of the selected architecture. AgentNAS-LLM is the seed architecture from Phase~1. AgentNAS-Full is the Phase~3 architecture with the best validation metric, excluding the seed itself: because the Phase~3 population contains the seed, a run in which NAS never beats it would otherwise report the seed again, hiding that NAS contributed nothing.

\subsection{Main Results}
\label{sec:results}
\begin{table}[t!]
\centering
\caption{NAS-Bench-360 results (lower is better), single run with random seed 42. AgentNAS-LLM is the seed architecture from Phase~1; AgentNAS-Full is the complete pipeline. \textbf{Best} and \underline{second best} per row.}
\label{tab:nas360_agentnas}
\resizebox{0.95\linewidth}{!}{
\begin{tabular}{l|c|rrrr|rr}
\toprule
Task & Metric & WRN16-4 & \makecell{GAEA\\DARTS} & Expert & \makecell{einspace\\(WRN)} & \makecell{AgentNAS\\-LLM} & \makecell{AgentNAS\\-Full} \\
\midrule
CIFAR-100  & $1 - \text{acc}$. (\%) & 23.35 & 24.02 & 19.39 & 21.47 & \underline{16.74} & \textbf{15.50} \\
Spherical  & $1 - \text{acc}$. (\%) & 85.77 & 48.23 & 67.41 & 66.37 & \underline{45.08} & \textbf{38.71} \\
NinaPro    & $1 - \text{acc}$. (\%) & \underline{6.78} & 17.67 & 8.73 & \textbf{6.37} & 8.35 & 8.04 \\
FSD50K     & $1-$mAP               & 0.92 & 0.94 & 0.62 & 0.65 & \underline{0.56} & \textbf{0.53} \\
Darcy Flow & Rel. L2               & 0.073 & 0.026 & 0.008 & 0.014 & \textbf{0.005} & \underline{0.006} \\
PSICOV     & MAE$_8$               & 3.84 & \textbf{2.94} & 3.35 & 4.38 & 3.37 & \underline{3.27} \\
Cosmic     & $1-$AUROC             & 0.245 & 0.229 & 0.127 & 0.730 & \underline{0.065} & \textbf{0.051} \\
ECG        & $1-$F1                & 0.43 & 0.34 & \underline{0.28} & 0.46 & 0.31 & \textbf{0.27} \\
Satellite  & $1 - \text{acc}$. (\%) & 15.49 & 12.51 & 19.80 & 12.55 & \underline{11.11} & \textbf{10.85} \\
DeepSea    & $1-$AUROC             & 0.40 & 0.36 & 0.30 & 0.36 & \textbf{0.26} & \underline{0.28} \\
\midrule
Average rank \hspace{0.5mm}$\downarrow$ & & 5.00 & 4.25 & 3.60 & 4.35 & \underline{2.30} & \textbf{1.50} \\
\bottomrule
\end{tabular}
}
\end{table}
\begin{table}[t!]
\centering
\caption{Unseen NAS test errors (\%, lower is better), single run with random seed 42. NAS~Best$^{\dagger}$: per-task best among six conventional NAS methods; einspace~Best$^{\ddagger}$: per-task best among all einspace variants. Full per-method results in Appendix~\ref{app:baselines}. \textbf{Best} and \underline{second best} per row.}
\label{tab:unseen_agentnas}
\begin{tabular}{l|r|rr|rr}
\toprule
Task & RN18 & \makecell{NAS\\Best$^{\dagger}$} & \makecell{einspace\\Best$^{\ddagger}$} & \makecell{AgentNAS\\-LLM} & \makecell{AgentNAS\\-Full} \\
\midrule
AddNIST     & 6.64  & \textbf{2.09}     & \underline{2.28}  & 2.82              & 2.42 \\
Language    & 7.84  & 7.57              & 2.08              & \underline{0.85}  & \textbf{0.49} \\
MultNIST    & 8.64  & \textbf{1.90}     & 3.63              & 3.54  & \underline{3.53} \\
CIFARTile   & 52.87 & \textbf{7.72}     & 37.24             & 11.20 & \underline{8.48} \\
Gutenberg   & 56.68 & 50.88             & 45.98 & \underline{44.63}    & \textbf{44.17} \\
GeoClassing & 9.92  & 3.97  & 4.69              & \underline{1.60}     & \textbf{0.97} \\
Chesseract  & 40.65 & \textbf{31.17}    & 38.14             & 33.84             & \underline{31.39} \\
\midrule
Average error \hspace{0.5mm}$\downarrow$ & 26.18 & 15.04 & 19.15 & \underline{14.07} & \textbf{13.06} \\
Average rank  \hspace{0.5mm}$\downarrow$ & 5.00 & \underline{2.14} & 3.43 & 2.71 & \textbf{1.71} \\
\bottomrule
\end{tabular}
\end{table}

Tables~\ref{tab:nas360_agentnas} and~\ref{tab:unseen_agentnas} show the results. AgentNAS gives the best result on 11 of the 17 tasks. On NAS-Bench-360, it is best on 8 of 10 tasks, with an average rank of 1.50 against 3.60 for the next-best baseline (Expert). On Unseen NAS, where the LLM works blind, AgentNAS-Full has an average error of 13.06\%, against 15.04\% for the per-task best conventional NAS method.

The seed from Phase~1 alone already outperforms the best published baseline on 10 tasks, without any predefined search space. AgentNAS-Full improves on the seed on a majority of tasks, with the largest gain on Spherical ($45.08\!\to\!38.71$), marginal gains on some tasks (e.g., MultNIST, $3.54\!\to\!3.53$), and no improvement on Darcy Flow and DeepSea.

\section{Ablation Studies}
\label{sec:ablation}

The following ablations use six tasks chosen to cover both benchmark suites, several modalities, and the full range of NAS gains in Tables~\ref{tab:nas360_agentnas} and~\ref{tab:unseen_agentnas}: large gains (Spherical, CIFARTile, Chesseract), small gains (NinaPro, Language), and no gain (Darcy Flow). The ablation studies demonstrate that the LLM improves the architecture rapidly within a small budget, but it also saturates earlier than NAS.

\subsection{LLM Seed Search vs.\ Fixed-Seed NAS}
\label{sec:abl-seed}

\begin{table}[t!]
\centering
\caption{Fixed-seed ablation (lower is better). To separate the contribution of the LLM seed from that of the slotted search, we replace the Phase~1 seed with a fixed published architecture (WRN16-4 on NAS-Bench-360, ResNet-18 on Unseen NAS). From left to right:
\emph{Reported} and \emph{Reproduced} are the fixed seed trained without any search, as reported by \citet{ericsson2024einspace} and as reproduced by us (the two differ, as einspace also observed for WRN16-4 on NAS-Bench-360);
\emph{einspace} searches 500 architectures starting from the fixed seed;
\emph{Phases~2--3} build a slotted search space around the fixed seed and search it with 160 evaluations;
\emph{AgentNAS-LLM} is the Phase~1 seed after 160 LLM evaluations, i.e., the same budget as Phases~2--3.
On Darcy Flow, WRN16-4 exceeds the 11\,GB VRAM limit, so \emph{Reproduced} shows the published value from Table~\ref{tab:nas360_agentnas} and Phases~2--3 could not be run (OOM).
Per-task metrics are as in Tables~\ref{tab:nas360_agentnas} and~\ref{tab:unseen_agentnas}. \textbf{Best} and \underline{second best} per row.}
\label{tab:agentnas_fixedseed}
\resizebox{0.8\textwidth}{!}{%
\begin{tabular}{l|rr|rr|r}
\toprule
& \multicolumn{2}{c|}{Fixed seed, no search} & \multicolumn{2}{c|}{Search from fixed seed} &  \\
Task & Reported & Reproduced & \makecell{einspace\\(500 evals)} & \makecell{Phases 2--3\\(160 evals)} & \makecell{AgentNAS-LLM\\(160 evals)} \\
\midrule
Spherical   & 76.32 & 71.65 & 66.37             & \underline{53.51} & \textbf{45.08} \\
NinaPro     & 10.32 & 7.13  & \textbf{6.37}     & \underline{6.68}  & 8.35  \\
Darcy Flow  & 0.032 & 0.073 & \underline{0.014} & \textit{OOM}      & \textbf{0.005} \\
Language    & 7.84  & 7.40  & \underline{3.16}  & 3.35              & \textbf{0.85}  \\
CIFARTile   & 55.88 & 52.87 & 39.35             & \underline{21.99} & \textbf{11.20} \\
Chesseract  & 44.46 & 40.65 & \underline{39.69} & 40.19             & \textbf{33.84} \\
\bottomrule
\end{tabular}}
\end{table}

AgentNAS combines two sources of strength: the LLM produces a high-quality seed (Phase~1), and NAS refines it within an automatically constructed search space (Phases~2--3).
To disentangle these contributions, we compare three configurations in Table~\ref{tab:agentnas_fixedseed}: (i)~a fixed published architecture (WideResNet-16-4 or ResNet-18) refined by Phases~2--3, (ii)~the same architecture refined by einspace~\citep{ericsson2024einspace} with regularized evolution over 500~evaluations, and (iii)~the full AgentNAS pipeline.

\paragraph{LLM seed search is more budget-efficient than fixed-seed NAS.}
AgentNAS-LLM outperforms the fixed-seed variant on every completed task except for Ninapro, often by a wide margin. This confirms that the LLM's ability to design a task-adapted seed provides a stronger starting point than beginning NAS from a generic architecture.

\paragraph{Phase~2 still constructs a strong search space.}
Even with a fixed seed, Phases~2--3 with 160~evaluations outperforms einspace with 500~evaluations on Spherical and CIFARTile. This suggests that the LLM-constructed slotted space is itself more effective than the grammar-based space of einspace, independent of seed quality. Combined, these two results justify the pipeline's design: the LLM contributes a strong seed \emph{and} a strong search space, and NAS exploits the latter to improve the former.

\begin{table}[t!]
\centering
\caption{Architecture versus training recipe (test metric, lower is better).
(a)~\emph{First arch}: the first architecture successfully evaluated in Phase~1, fully trained;
\emph{Recipe-only}: the same architecture after 160 evaluations of recipe-only tuning;
\emph{AgentNAS-LLM}: 160 evaluations of joint architecture and recipe search.
(b)~AgentNAS-Full retrained with the fixed recipe of einspace (Table~\ref{tab:hyperparam_comparison}), compared with einspace, which uses RE from WRN16-4 on NAS-Bench-360 and from a mix of seeds on Unseen NAS.
AgentNAS-Full with our recipe is shown for reference.
\textbf{Bold}: better of \emph{Recipe-only} and \emph{AgentNAS-LLM} in~(a), and of the two methods in~(b).
Metrics as in Tables~\ref{tab:nas360_agentnas} and~\ref{tab:unseen_agentnas}.}
\label{tab:recipe_control}
\resizebox{0.97\linewidth}{!}{%
\begin{tabular}{l|rrr|rr|r}
\toprule
& \multicolumn{3}{c|}{(a) Recipe vs.\ architecture} & \multicolumn{2}{c|}{(b) Using einspace's recipe} & Reference \\
Task & First arch & Recipe-only & AgentNAS-LLM & einspace & AgentNAS-Full & \makecell{AgentNAS-Full} \\
\midrule
Spherical  & 77.32 & 68.41          & \textbf{45.08} & 66.37          & \textbf{58.83} & 38.71 \\
NinaPro    & 13.81 & 10.32          & \textbf{8.35}  & \textbf{6.37}  & 8.50           & 8.04  \\
Darcy Flow & 0.022 & 0.019          & \textbf{0.005} & \textbf{0.014} & 0.017          & 0.006 \\
Language   & 0.32  & \textbf{0.27}  & 0.85           & 2.08           & \textbf{1.81}  & 0.49  \\
CIFARTile  & 69.64 & 57.23          & \textbf{11.20} & 37.24          & \textbf{17.99} & 8.48  \\
Chesseract & 44.57 & 40.93          & \textbf{33.84} & 38.14          & \textbf{30.79} & 31.39 \\
\bottomrule
\end{tabular}}
\end{table}

\subsection{Architecture vs. Training Recipe}
The LLM chooses the training recipe together with the architecture, and Phase~3 also searches the learning rate, so part of the gain could come from the recipe rather than the architecture. Table~\ref{tab:recipe_control} separates the two in two ways. First, we fix the first architecture that the LLM proposes in Phase~1 and let it tune only the recipe for 160 evaluations. Recipe tuning improves this architecture on every task, but architecture search improves it far more on five of the six tasks. The exception is Language, where the first architecture already has the lowest test error even though Phase~1 reduces the proxy validation error from 9.98\% to 5.98\%; we return to this mismatch between validation and test in Section~\ref{sec:conclusion}. Second, we retrain the AgentNAS-Full architectures with the fixed recipe of einspace. Under this recipe, they still outperform einspace on four of the six tasks (Spherical, Language, CIFARTile, and Chesseract), so the architecture alone accounts for much of the gain over einspace. On Darcy Flow, where einspace trains for 200 epochs and our recipe for 1{,}000, the training length likely explains part of the remaining gap.

\subsection{Complementarity of LLM and NAS}
\label{sec:abl-complement}

\begin{figure*}[t]
  \centering
  \includegraphics[width=\linewidth]{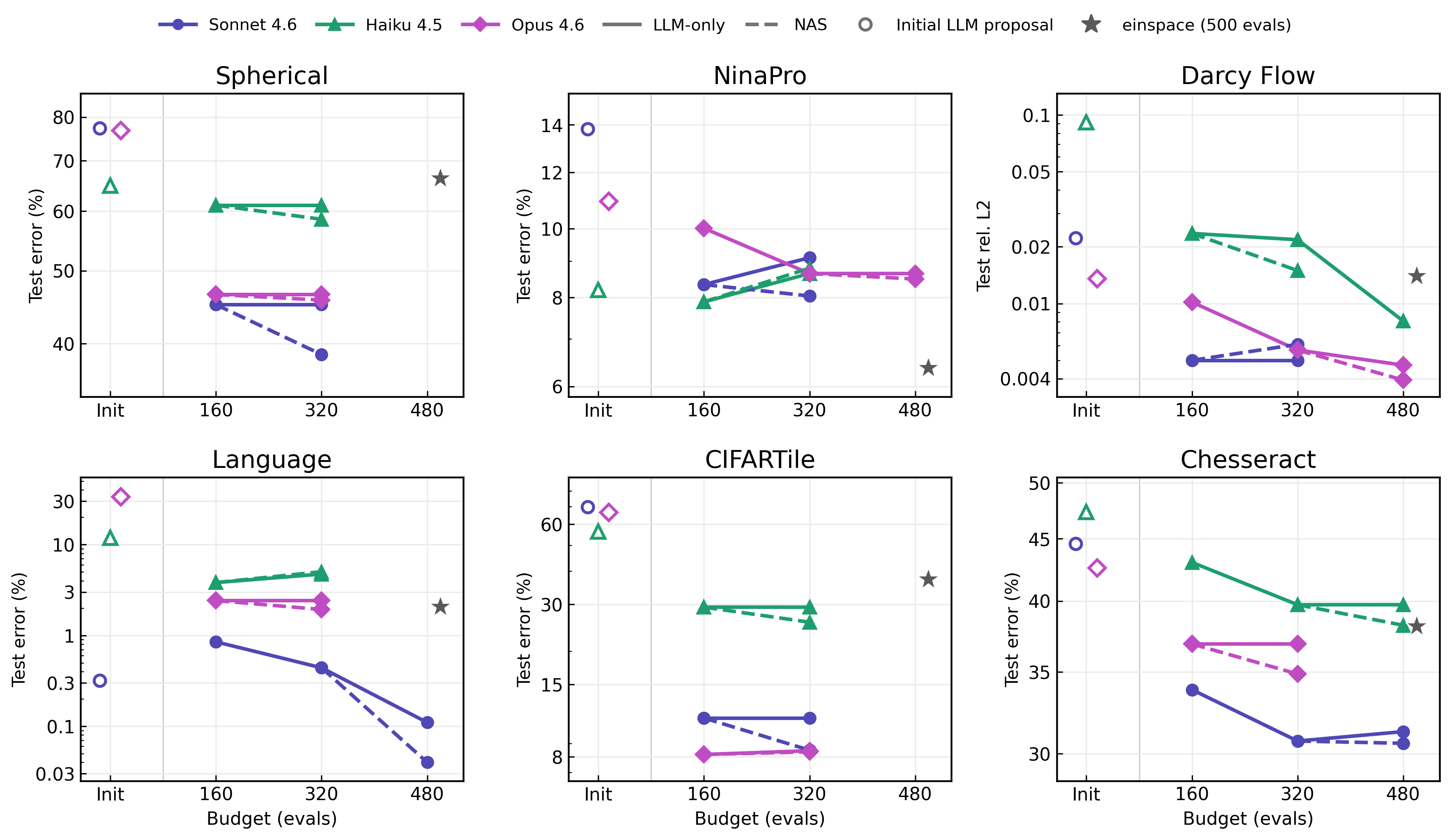}
  \caption{Budget--performance trajectories (log scale, lower is better). \textbf{Solid}: LLM-only search; \textbf{dashed}: NAS launched from the point where LLM-only search saturates. Hollow markers (\emph{Init}) show the first architecture successfully evaluated in Phase~1, fully trained. Each point is the test metric of the architecture selected by validation, so trajectories need not decrease monotonically. Stars show einspace at its own budget of 500 evaluations (RE from WRN16-4 on NAS-Bench-360 and from a mix of seeds on Unseen NAS~\citep{ericsson2024einspace}), plotted in a separate column. On Darcy Flow, the search spaces Haiku built from its 320-budget seed failed our validation check in repeated trials, so NAS starts from its 160-budget seed instead.}
  \label{fig:ablation}
\end{figure*}

An LLM proposes each architecture as a new sample conditioned on past results rather than a systematic recombination, whereas NAS recombines modules across slots. We call the two complementary on a task if, for a fixed total budget, switching from LLM sampling to NAS once LLM sampling stops improving yields a better final architecture than spending the whole budget on LLM sampling. Figure~\ref{fig:ablation} tests this with three LLMs. Starting from 160 LLM evaluations, we add 160 more LLM evaluations as long as the validation metric keeps improving, up to 480 in total, and run 160 evaluations of NAS at the same point. Under the same total budget, switching to NAS outperforms LLM-only search on several task--model pairs, continuing to improve beyond the point where LLM-only search saturates.

\subsection{Robustness}

\begin{table}[t!]
\centering
\caption{Ten random seeds (42--51) on the six ablation tasks (test metric, lower is better; errors in \%, Darcy Flow in relative L2 $\times10^{-3}$). LLM: AgentNAS-LLM; Full: AgentNAS-Full. Std is the sample standard deviation over the ten seeds. $T$, $p$: one-sided Wilcoxon signed-rank test of the hypothesis that AgentNAS-Full improves on AgentNAS-LLM, with $T=\min(W^{+},W^{-})$; $r$: rank-biserial correlation, positive when NAS improves. On NinaPro, one seed with identical test error in phase 1 and phase 3 architectures is discarded as a zero difference ($n=9$). Per-task metrics are as in Tables~\ref{tab:nas360_agentnas} and~\ref{tab:unseen_agentnas}.}
\label{tab:multiseed}
\resizebox{0.97\linewidth}{!}{%
\begin{tabular}{l|cc|cc|rrr}
\toprule
& \multicolumn{2}{c|}{Mean $\pm$ std} & \multicolumn{2}{c|}{Median (IQR)} & & & \\
Task & LLM & Full & LLM & Full & $T$ & $p$ & $r$ \\
\midrule
Spherical  & 47.42 $\pm$ 4.01 & 45.21 $\pm$ 3.93 & 46.67 (43.20--51.13) & 45.43 (39.86--49.43) & 6  & \textbf{.014} & .78 \\
CIFARTile  & 10.34 $\pm$ 2.93 & 9.09 $\pm$ 3.35  & 10.46 (8.47--11.47)  & 8.35 (6.29--11.06)   & 10 & \textbf{.042} & .64 \\
Chesseract & 36.59 $\pm$ 3.14 & 34.59 $\pm$ 2.82 & 36.85 (34.40--38.83) & 35.10 (31.88--36.37) & 3  & \textbf{.005} & .89 \\
NinaPro    & 8.10 $\pm$ 1.63  & 8.21 $\pm$ 1.46  & 8.04 (7.47--8.46)    & 8.19 (7.74--9.06)    & 20 & .633 & $-$.11 \\
Darcy Flow & 5.73 $\pm$ 1.27  & 6.57 $\pm$ 1.99  & 5.90 (4.83--6.26)    & 6.39 (5.90--6.43)    & 18 & .839 & $-$.35 \\
Language   & 1.64 $\pm$ 1.31  & 2.10 $\pm$ 1.50  & 1.16 (0.87--2.88)    & 1.83 (1.18--3.02)    & 17 & .862 & $-$.38 \\
\bottomrule
\end{tabular}}
\end{table}

\paragraph{Variation across random seeds.}
Table~\ref{tab:multiseed} repeats the full pipeline with ten random seeds on the six ablation tasks and tests whether AgentNAS-Full improves on AgentNAS-LLM with a one-sided Wilcoxon signed-rank test. The improvement is significant on Spherical, CIFARTile, and Chesseract, with large effect sizes ($r\geq0.64$). On NinaPro and Language, NAS improves the median validation error (NinaPro $7.44\!\to\!7.36$, Language $7.47\!\to\!6.92$) but not the test error, and on Darcy Flow it improves neither. The seed-42 results in Tables~\ref{tab:nas360_agentnas} and~\ref{tab:unseen_agentnas} lie within these distributions, though on Spherical the seed-42 result is among the best of the ten. Averaged over seeds, AgentNAS-Full remains ahead of the best published baseline on Spherical (45.21\% vs.\ 48.23\%) and Darcy Flow (0.0066 vs.\ 0.008), and is within 0.02 percentage points of it on Language (2.10\% vs.\ 2.08\%).

\paragraph{Search algorithm.}
Within the same slotted space, regularized evolution and random search perform comparably, each winning on three of the six tasks, while GDAS is much worse on Darcy Flow and CIFARTile with untuned hyperparameters (Appendix~\ref{app:search_algorithm}). The search space matters more than the search algorithm: random search with 160 evaluations in the slotted space reaches 0.46\%, 11.28\%, and 33.51\% on Language, CIFARTile, and Chesseract, against 12.99\%, 69.10\%, and 38.54\% for random search with 1{,}000 evaluations in the einspace's search space.

\paragraph{Other LLM families.}
With GPT-5.4 and Qwen3.6-35B-A3B on the six ablation tasks (Appendix~\ref{app:llm_families}), Phase~1 produces competitive seeds with every model; Qwen3.6 even finds stronger seeds than both GPT-5.4 and Sonnet on Spherical and CIFARTile. NAS improves the GPT-5.4 seeds on Spherical, CIFARTile, and Chesseract, the same three tasks on which it significantly improves the Sonnet seeds (Table~\ref{tab:multiseed}). With Qwen3.6, NAS improves only Chesseract and is worse on the other five tasks, most sharply on Language and CIFARTile, where the Slotted Architecture Planner produced defective slot alternatives. Together with the Haiku failure on Darcy Flow, this suggests that Phase~1 works across a wide range of models, whereas Phase~2 requires a minimum level of model capability.

\subsection{What Does NAS Change?}
\label{sec:abl-qualitative}

We now look at how NAS changes the seed, using the architectures from Figure~\ref{fig:ablation}.

\paragraph{Chesseract: NAS adds what the seed lacks.}
Chesseract is a three-class task (white wins, draw, black wins) on $8\times8$ boards encoded as 12 channels (2 players $\times$ 6 piece types). The three LLMs produce very different seeds: a chess-specific network with one branch per piece type (Sonnet), a generic Transformer (Opus), and a CNN (Haiku). In all three cases, NAS adds capabilities the seed lacks and leaves effective components in place. The Transformer and CNN seeds have no gating, and NAS adds it through ReZero scaling, sigmoid gates, and gated feed-forward layers (GLU, SwiGLU). The Sonnet seed already has a sigmoid gate per piece-type branch; NAS removes these and places a single gate where the branches merge. NAS also adds depthwise $3\times3$ convolutions to the Transformer seed, which otherwise has no local mixing over the board, but does not add global attention to the CNN seed. In stages where they contribute little, it removes skip connections or attention layers.

\paragraph{CIFARTile: channel attention.}
CIFARTile presents 2$\times$2 tiled CIFAR-10 images as a 4-class classification task. Across three different types of backbones (RegNet, ResNet, Bottleneck variants), NAS consistently strengthens channel attention: adding SE blocks where none existed, or layering complementary mechanisms (Coordinate Attention, ECA) on top of existing SE modules. This pattern suggests that channel recalibration is uniformly beneficial for this task, where discriminative signals may be concentrated in specific spatial tiles.

\paragraph{Darcy Flow: the search space decides.}
All three LLMs propose Fourier Neural Operator seeds for Darcy Flow, but NAS improves only the Opus seed. In the Sonnet search space, every alternative for the spectral layers keeps the seed's channel-mixing spectral convolution, so NAS has nothing meaningful to choose there. The Opus search space includes an alternative that scales each channel's Fourier modes independently with a lightweight weight and leaves channel mixing to $1\times1$ convolutions, and NAS selects it. Whether NAS helps can therefore depend on the alternatives the LLM puts into the slots, not only on the quality of the seed.

\section{Conclusion}
\label{sec:conclusion}
We proposed the slotted architecture, through which an LLM turns its own seed architecture into a task-specific search space for conventional NAS, and implemented it in AgentNAS, which outperforms published baselines on 11 of 17 tasks. Our ablations suggest a clear division of labor: LLM sampling reaches a strong architecture quickly but saturates, and switching to NAS at that point outperforms further LLM sampling at the same budget.

\paragraph{Limitations.}
Our evaluation is subject to several constraints. Our benchmarks consist of small- to medium-scale tasks on which convolutional networks are competitive, and we do not test large-scale or generative tasks (Appendix~\ref{app:families}). The main tables report one run per task, and only six tasks are repeated over ten seeds, on three of which NAS gives a significant gain. Selection relies on validation metrics, which do not always transfer to test, as on DeepSea, NinaPro, and Language. Each candidate trains on a single RTX2080Ti, which has low VRAM of 11\,GB and low throughput, potentially favoring small architectures. (Appendix~\ref{app:efficiency}). Early stopping makes the number of evaluations actually used vary across runs, and baseline results come from their original papers with different protocols. Finally, Phase~2 requires a minimum level of LLM capability, as the failures of Haiku and Qwen3.6 show.

\paragraph{Future work.}
Three directions are most promising. The question of what makes a good LLM-constructed search space and how to systematically improve it remains largely unexplored. Additionally, our search algorithms (regularized evolution, GDAS) and their hyperparameters are not optimized; exploring alternatives, especially efficient weight-sharing methods, within the slotted framework is a natural next step. Finally, extending the pipeline to generative tasks would further test its generality, though this likely requires relaxing the current hardware and time constraints.

\begin{ack}
This work was supported by Institute of Information \& Communications Technology Planning \&
Evaluation (IITP) grant funded by the Korea government (MSIT) (No. IITP-2026-RS-2024-00360227,
Leading Generative AI Human Resources Development, No. RS-2025-25442824, AI Star Fellowship Program (Ulsan National Institute of Science and Technology), \& No. RS-2020-II201336, Artificial Intelligence graduate school support (UNIST)). 
This work was also supported by the National Research Foundation of Korea (NRF) grants funded by the Korea government (MSIT) (No. RS-2025-24913002).
\end{ack}

\bibliographystyle{plainnat}
\bibliography{references}

\newpage


\appendix

\section{Efficiency Analysis}
\label{app:efficiency}

We report search cost, parameter counts, and FLOPs for all tasks across both benchmarks. Tables~\ref{tab:nb360_comparison} and~\ref{tab:nb360_transparency} cover NAS-Bench-360; Tables~\ref{tab:unseen_nas_comparison} and~\ref{tab:unseen_nas_transparency} cover Unseen NAS. We separate \emph{comparison} tables (including baselines where available) from \emph{transparency} tables (our method only, for metrics where no baseline data exists).

\paragraph{Search cost.}
AgentNAS is more expensive than supernet-based methods such as GAEA, which amortize search cost across a shared network. This is expected: our pipeline trains each candidate individually through regularized evolution, as einspace. Compared to einspace, which is the most directly comparable baseline, AgentNAS is in a similar range on Unseen NAS tasks.

\paragraph{Parameter counts.}
In Unseen NAS, AgentNAS yields parameter counts much smaller than einspace in general, and even comparable to those of conventional NAS methods. In NAS-Bench-360, AgentNAS architectures tend to be larger than GAEA and expert-designed baselines. This is most pronounced on tasks where the LLM designs deep multi-stage networks (e.g., CIFAR-100: 23.4M vs.\ 4.9M for GAEA; Satellite: 67.3M vs.\ 3.4M). Expert architectures are typically the most parameter-efficient, reflecting the benefit of domain-specific engineering. Notably, Phase~3 tends to reduce parameter count relative to the Phase~1 seed (e.g., Spherical: 15.4M vs.\ 19.7M; Satellite: 54.7M vs.\ 67.3M), indicating that NAS can tighten the architecture as well as expand it.

\paragraph{FLOPs.}
Despite larger parameter counts, AgentNAS architectures are frequently more compute-efficient than baselines in terms of FLOPs. On NAS-Bench-360, our architectures use fewer FLOPs than GAEA on 8 of 10~tasks. This pattern implies that the AgentNAS search space may generally favor parameter-heavy but compute-efficient operations, such as depthwise separable convolutions or mobile inverted bottleneck blocks.

\paragraph{Memory and time constraints.}
Of the roughly 4{,}500 architectures evaluated in the main experiments, only 1.5\% ran out of memory. Successful architectures used a median of 27\% of the 11\,GB VRAM, and none exceeded 95\%. In contrast, 37.6\% of successful proxy runs used at least 95\% of the one-hour time cap. The binding constraint during the search is therefore training time rather than memory. The LLM does take memory into account, but mostly as a trade-off between width and depth rather than by avoiding particular topologies.

\section{Cost}
\label{app:cost}
 
Table~\ref{tab:cost} gives the cost of the main run on every task. LLM cost is computed from token counts at \$3 per million input tokens and \$15 per million output tokens (Claude Sonnet~4.6). GPU cost is the wall-clock time of the whole pipeline, including Phases~1--3 and all full-training runs (Tables~\ref{tab:unseen_nas_comparison} and~\ref{tab:nb360_transparency}), multiplied by \$0.649 per hour, the rental price of an eight-GPU RTX~2080\,Ti node on Vast.ai. The total over the 17 main runs is \$988, of which \$220 is LLM cost; ablations and repeated seeds are not included.
 
Phase~1 accounts for 84\% of the LLM cost, because the Planner and the Code Generator are called repeatedly over up to 160 evaluations; Phase~2 costs \$2 per task on average. LLM cost scales with the length of the architecture code kept in context rather than with model size. Chesseract is the most expensive task (\$33.69) despite having only about 2M parameters, because its chess-specific architectures average about 2{,}500 tokens; Satellite costs \$4.77 with 67M parameters, as its architectures average about 470 tokens. GPU cost dominates on most tasks and depends mainly on the training time per candidate.

\begin{table}[h]
    \caption{Wall-clock runtime and parameter counts on Unseen NAS tasks. Reference methods use internal GPU clusters~\citep{geada2024insights}; our experiments use $8\times$RTX\,2080\,Ti. RE(\texttt{einspace}) is the RE(Mix) variant; RE(hNB201) searches from scratch. PC-DARTS entries are missing where logs were unavailable, which is denoted by \citet{ericsson2024einspace}.}
    \label{tab:unseen_nas_comparison}
    \setlength\extrarowheight{2pt}
    \centering
    \resizebox{\linewidth}{!}{%
    \begin{tabular}{lrrrrrrr}
    \toprule
              & AddNIST & Language & MultNIST & CIFARTile & Gutenberg & GeoClassing & Chesseract \\
    \midrule
    \multicolumn{8}{l}{\textit{Wall-clock (hours)}} \\[2pt]
    DrNAS              & 10  & 9   & 11  & 25   & 13  & 23  & 10  \\
    PC-DARTS           & 4   & --  & 5   & 12   & 9   & --  & 2   \\
    RE(hNB201)         & 15  & 72  & 21  & 59   & 9   & 37  & 9   \\
    RE(\texttt{einspace}) & 55  & 71  & 32  & 62   & 42  & 65  & 42  \\
    \cmidrule{1-8}
    Ours (P1)          & 67  & 5   & 69  & 86   & 4   & 67  & 59  \\
    Ours (P3)          & 11  & 2   & 19  & 30   & 2   & 19  & 11  \\
    Ours (P1{+}P3)     & 78  & 7   & 87  & 116  & 7   & 86  & 69  \\
    \midrule
    \multicolumn{8}{l}{\textit{\#Params ($\times10^6$)}} \\[2pt]
    DrNAS              & 4   & 4   & 5   & 3    & 3   & 4   & 4   \\
    PC-DARTS           & 3   & --  & 3   & 3    & 2   & --  & 2   \\
    RE(hNB201)         & 1   & 1   & 1   & 3    & 1   & 1   & 1   \\
    RE(\texttt{einspace}) & 20  & 1   & 25  & 5    & 1   & 4   & 11  \\
    \cmidrule{1-8}
    Ours (P1)          & 6   & 1   & 11  & 3    & 1   & 11  & 2   \\
    Ours (P3)          & 1   & 1   & 11  & 3    & 1   & 7   & 2   \\
    \bottomrule
    \end{tabular}%
    }
\end{table}


\begin{table}[h]
    \caption{GPU time and FLOPs of AgentNAS architectures on Unseen NAS tasks. No reference data is available for these metrics. Experiments use $8\times$RTX\,2080\,Ti.}
    \label{tab:unseen_nas_transparency}
    \setlength\extrarowheight{2pt}
    \centering
    \resizebox{\linewidth}{!}{%
    \begin{tabular}{lrrrrrrr}
    \toprule
              & AddNIST & Language & MultNIST & CIFARTile & Gutenberg & GeoClassing & Chesseract \\
    \midrule
    \multicolumn{8}{l}{\textit{GPU time (hours)}} \\[2pt]
    Ours (P1)          & 103 & 14  & 144 & 165  & 14  & 125 & 40  \\
    Ours (P3)          & 67  & 14  & 135 & 202  & 12  & 114 & 65  \\
    Ours (P1{+}P3)     & 170 & 28  & 278 & 367  & 26  & 239 & 105 \\
    \midrule
    \multicolumn{8}{l}{\textit{FLOPs ($\times10^6$)}} \\[2pt]
    Ours (P1)          & 223  & 17   & 1400  & 204   & 12   & 2020  & 592  \\
    Ours (P3)          & 151  & 18   & 1400  & 280   & 4    & 1170  & 1130 \\
    \bottomrule
    \end{tabular}%
    }
\end{table}


\begin{table}[H]
\caption{GPU time, parameter counts, and FLOPs on NAS-Bench-360 tasks. GAEA values are from~\citet{tu2022bench} using a single V100 16\,GB; params and FLOPs are means over 3~seeds. Our experiments use $8\times$RTX\,2080\,Ti.}
    \label{tab:nb360_comparison}
    \setlength\extrarowheight{2pt}
    \centering
    \resizebox{\linewidth}{!}{%
    \begin{tabular}{lrrrrrrrrrr}
    \toprule
              & CIFAR-100 & Spherical & Darcy Flow & PSICOV & Cosmic & NinaPro & FSD50K & ECG & Satellite & DeepSEA \\
    \midrule
    \multicolumn{11}{l}{\textit{GPU time (hours)}} \\[2pt]
    GAEA               & 9.5  & 16.5 & 6.5  & 18    & 21.5 & 0.5  & 37   & 140  & 28   & 39.5 \\
    \cmidrule{1-11}
    Ours (P1)          & 99   & 125  & 111  & 199   & 43   & 4    & 258  & 316  & 42   & 294  \\
    Ours (P3)          & 186  & 228  & 121  & 191   & 51   & 5    & 240  & 383  & 106  & 399  \\
    Ours (P1{+}P3)     & 285  & 353  & 233  & 390   & 94   & 9    & 498  & 698  & 148  & 693  \\
    \midrule
    \multicolumn{11}{l}{\textit{\#Params ($\times10^6$)}} \\[2pt]
    GAEA               & 4.9  & 1.7  & 0.6  & 0.5   & 0.4  & 3.4  & 0.8  & 3.3  & 3.4  & 2.9  \\
    Expert             & 3.1  & 0.16 & 1.2  & 0.6   & 0.10 & 1.4  & 0.35 & 16.5 & 0.48 & 60.9 \\
    \cmidrule{1-11}
    Ours (P1)          & 23.4 & 19.7 & 1.8  & 1.0   & 12.9 & 1.2  & 10.3 & 0.64 & 67.3 & 35.8 \\
    Ours (P3)          & 26.0 & 15.4 & 1.2  & 0.59 & 11.2 & 1.2  & 10.3 & 0.24 & 54.7 & 25.0 \\
    \midrule
    \multicolumn{11}{l}{\textit{FLOPs ($\times10^9$)}} \\[2pt]
    GAEA               & 1.42 & 1.91 & 9.33 & 17.74 & 14.27& 0.89 & 2.57 & 2.28 & 0.11 & 2.01 \\
    Expert             & 1.18 & n/a\textsuperscript{*}  & n/a\textsuperscript{*}  & 0.01  & 1.96 & 0.02 & 0.66 & 0.70 & 0.01 & 0.12 \\
    \cmidrule{1-11}
    Ours (P1)          & 3.36 & 0.88 & 3.65 & 16.64 & 8.21 & 0.23 & 0.57 & 0.63 & 0.07 & 5.71 \\
    Ours (P3)          & 3.51 & 0.83 & 1.86 & 15.6 & 8.43 & 0.19 & 0.59 & 0.49 & 0.05 & 5.04 \\
    \bottomrule
    \multicolumn{11}{l}{\footnotesize \textsuperscript{*} Expert models contain non-standard modules without FLOPs count~\citep{tu2022bench}.}
    \end{tabular}%
    }
\end{table}


\begin{table}[H]
    \caption{Wall-clock runtime of AgentNAS on NAS-Bench-360 tasks. No wall-clock reference is available for GAEA. Experiments use $8\times$RTX\,2080\,Ti.}
    \label{tab:nb360_transparency}
    \setlength\extrarowheight{2pt}
    \centering
    \resizebox{\linewidth}{!}{%
    \begin{tabular}{lrrrrrrrrrr}
    \toprule
              & CIFAR-100 & Spherical & Darcy Flow & PSICOV & Cosmic & NinaPro & FSD50K & ECG & Satellite & DeepSEA \\
    \midrule
    \multicolumn{11}{l}{\textit{Wall-clock (hours)}} \\[2pt]
    Ours (P1)          & 15   & 24   & 19   & 60    & 39   & 2    & 66   & 80   & 9    & 89   \\
    Ours (P3)          & 27   & 30   & 18   & 56    & 33   & 1    & 35   & 63   & 15   & 53   \\
    Ours (P1{+}P3)     & 43   & 54   & 36   & 115   & 73   & 3    & 101  & 143  & 23   & 142  \\
    \bottomrule
    \end{tabular}%
    }
\end{table}

\begin{table}[H]
\centering
\caption{Cost of the main run per task. Tokens in millions. LLM cost at \$3/\$15 per million input/output tokens; GPU cost is wall-clock hours $\times$ \$0.649 per hour for an eight-GPU RTX~2080\,Ti node.}
\label{tab:cost}
\resizebox{\linewidth}{!}{%
\begin{tabular}{l|rrr|rrr|r|r|rrr|r|r}
\toprule
& \multicolumn{3}{c|}{Phase~1 tokens} & \multicolumn{3}{c|}{Phase~2 tokens} & & Wall & \multicolumn{3}{c|}{LLM cost (\$)} & GPU & Total \\
Task & in & out & total & in & out & total & Calls & clock (h) & P1 & P2 & sum & cost (\$) & (\$) \\
\midrule
AddNIST     & 1.54 & 0.26 & 1.79 & 0.06 & 0.05 & 0.11 & 309 & 78  & 8.52  & 0.93 & 9.45  & 50.62 & 60.07 \\
Chesseract  & 6.00 & 0.93 & 6.93 & 0.08 & 0.10 & 0.17 & 388 & 69  & 31.95 & 1.74 & 33.69 & 44.78 & 78.47 \\
CIFAR-100   & 1.19 & 0.17 & 1.36 & 0.19 & 0.16 & 0.34 & 485 & 43  & 6.12  & 2.97 & 9.09  & 27.91 & 37.00 \\
CIFARTile   & 2.90 & 0.42 & 3.32 & 0.32 & 0.36 & 0.68 & 754 & 116 & 15.00 & 6.36 & 21.36 & 75.28 & 96.64 \\
Cosmic      & 1.66 & 0.28 & 1.95 & 0.07 & 0.09 & 0.16 & 296 & 73  & 9.18  & 1.56 & 10.74 & 47.38 & 58.12 \\
Darcy Flow  & 2.26 & 0.35 & 2.60 & 0.07 & 0.08 & 0.15 & 323 & 36  & 12.03 & 1.41 & 13.44 & 23.36 & 36.80 \\
DeepSEA     & 1.60 & 0.25 & 1.84 & 0.06 & 0.05 & 0.11 & 290 & 142 & 8.55  & 0.93 & 9.48  & 92.16 & 101.64 \\
ECG         & 1.44 & 0.24 & 1.68 & 0.05 & 0.06 & 0.12 & 278 & 143 & 7.92  & 1.05 & 8.97  & 92.81 & 101.78 \\
FSD50K      & 2.31 & 0.38 & 2.69 & 0.07 & 0.05 & 0.12 & 344 & 101 & 12.63 & 0.96 & 13.59 & 65.55 & 79.14 \\
GeoClassing & 1.64 & 0.29 & 1.93 & 0.26 & 0.26 & 0.52 & 664 & 86  & 9.27  & 4.68 & 13.95 & 55.81 & 69.76 \\
Gutenberg   & 1.60 & 0.22 & 1.81 & 0.10 & 0.07 & 0.17 & 321 & 7   & 8.10  & 1.35 & 9.45  & 4.54  & 13.99 \\
Language    & 2.02 & 0.31 & 2.33 & 0.06 & 0.04 & 0.11 & 313 & 7   & 10.71 & 0.78 & 11.49 & 4.54  & 16.03 \\
MultNIST    & 1.71 & 0.27 & 1.97 & 0.11 & 0.07 & 0.19 & 361 & 87  & 9.18  & 1.38 & 10.56 & 56.46 & 67.02 \\
NinaPro     & 1.52 & 0.27 & 1.78 & 0.06 & 0.06 & 0.11 & 286 & 3   & 8.61  & 1.08 & 9.69  & 1.95  & 11.64 \\
PSICOV      & 2.38 & 0.35 & 2.73 & 0.08 & 0.07 & 0.15 & 348 & 115 & 12.39 & 1.29 & 13.68 & 74.64 & 88.32 \\
Satellite   & 0.72 & 0.12 & 0.84 & 0.07 & 0.04 & 0.11 & 270 & 23  & 3.96  & 0.81 & 4.77  & 14.93 & 19.70 \\
Spherical   & 2.14 & 0.37 & 2.52 & 0.24 & 0.28 & 0.52 & 644 & 54  & 11.97 & 4.92 & 16.89 & 35.05 & 51.94 \\
\midrule
Mean        & 2.04  & 0.32 & 2.36  & 0.11 & 0.11 & 0.23 & 392.6 & 69.6 & 10.95 & 2.01 & 12.96 & 45.16 & 58.12 \\
Total       & 34.63 & 5.48 & 40.07 & 1.95 & 1.89 & 3.84 & 6674  & 1183 & 186.09 & 34.20 & 220.29 & 767.77 & 988.06 \\
\bottomrule
\end{tabular}}
\end{table}

\section{Parameter-Constrained Search}
\label{app:param_constrained}
 
AgentNAS architectures are often larger than those of GAEA DARTS on NAS-Bench-360 (Table~\ref{tab:nb360_comparison}). On the three tasks with the largest gap in parameter count, CIFAR-100, Cosmic, and Satellite, we rerun the pipeline with a ceiling equal to the parameter count of GAEA DARTS. The ceiling is stated in the Planner and Slotted Architecture Planner prompts, and any candidate above it is rejected before training.
 
\begin{table}[h]
\centering
\caption{AgentNAS with a parameter ceiling matched to GAEA DARTS (lower is better). Unconstrained: AgentNAS-Full from Table~\ref{tab:nas360_agentnas}.}
\label{tab:param_constrained}
\begin{tabular}{ll|rrr}
\toprule
Task & Method & Params (M) & FLOPs (G) & Metric \\
\midrule
CIFAR-100 (1 $-$ acc., \%) & GAEA DARTS              & 4.9  & 1.42  & 24.02 \\
                          & AgentNAS (constrained)   & 4.1  & 0.51  & 18.10 \\
                          & AgentNAS (unconstrained) & 26.0 & 3.51  & \textbf{15.50} \\
\midrule
Cosmic (1 $-$ AUROC)      & GAEA DARTS               & 0.4  & 14.27 & 0.229 \\
                          & AgentNAS (constrained)   & 0.29 & 10.17 & \textbf{0.014} \\
                          & AgentNAS (unconstrained) & 11.2 & 8.43  & 0.051 \\
\midrule
Satellite (1 $-$ acc., \%) & GAEA DARTS              & 3.4  & 0.11  & 12.51 \\
                          & AgentNAS (constrained)   & 3.3  & 0.003 & \textbf{9.35} \\
                          & AgentNAS (unconstrained) & 54.7 & 0.05  & 10.85 \\
\bottomrule
\end{tabular}
\end{table}
 
With the same parameter budget, AgentNAS outperforms GAEA DARTS on all three tasks. On Cosmic and Satellite, the constrained runs even outperform the unconstrained ones. Under a tight budget, the architectures rely more on task-specific processing than on capacity: a log transform of the input for Cosmic, Fourier features of the time series for Satellite, and attention modules with stochastic depth for CIFAR-100.

\section{Training Recipes Generated by LLMs}
\label{app:recipe}
In this section, we describe the training recipes defined by each run. NAS uses the same recipes, except for the learning rate. We compare the training recipes with those of einspace~\citep{ericsson2024einspace}.

\begin{table}[H]
    \caption{Comparison of hyperparameters: einspace's fixed recipe vs.\ our dynamically searched configurations. einspace uses a fixed recipe per task that does not change during the search, whereas ours tailors each hyperparameter per dataset. When NAS chose different learning rates, we denote the change with $\rightarrow$.}
    \label{tab:hyperparam_comparison}
    \centering
    \resizebox{\linewidth}{!}{%
    \begin{tabular}{l|cccc|cccc}
        \toprule
        & \multicolumn{4}{c|}{einspace (fixed)} & \multicolumn{4}{c}{Ours (LLM-searched)} \\
        \cmidrule(lr){2-5} \cmidrule(lr){6-9}
        Dataset & Epochs & Batch & Learning & Weight & Epochs & Batch & Learning & Weight \\
        & & size & rate & decay & & size & rate & decay \\
        \midrule
        AddNIST     & 64   & 256  & 0.04  & $3 \times 10^{-4}$ & 200  & 128  & $0.003 \rightarrow 0.012$  & 0.02 \\
        Language    & 64   & 256  & 0.04  & $3 \times 10^{-4}$ & 200  & 256  & $3\times10^{-4} \rightarrow 6\times10^{-4}$ & 0.01 \\
        MultNIST    & 64   & 256  & 0.04  & $3 \times 10^{-4}$ & 300  & 128  & 0.2            & $5 \times 10^{-4}$ \\
        CIFARTile   & 64   & 256  & 0.04  & $3 \times 10^{-4}$ & 300  & 128  & 0.001          & 0.05 \\
        Gutenberg   & 64   & 256  & 0.04  & $3 \times 10^{-4}$ & 150  & 256  & $3\times10^{-4} \rightarrow 7.5\times10^{-5}$ & $1 \times 10^{-4}$ \\
        GeoClassing & 64   & 256  & 0.04  & $3 \times 10^{-4}$ & 200  & 128  & 0.1            & $5 \times 10^{-4}$ \\
        Chesseract  & 64   & 256  & 0.04  & $3 \times 10^{-4}$ & 200  & 256  & $3\times10^{-4}$ & $5 \times 10^{-4}$ \\
        \midrule
        CIFAR100   & 200  & 128  & 0.1   & $5 \times 10^{-4}$ & 200  & 128  & 0.1            & $5 \times 10^{-4}$ \\
        Spherical  & 200  & 128  & 0.1   & $5 \times 10^{-4}$ & 200  & 128  & $0.001 \rightarrow 0.004$   & 0.05 \\
        NinaPro    & 200  & 128  & 0.1   & $5 \times 10^{-4}$ & 300  & 32   & 0.001          & 0.01 \\
        FSD50K     & 200  & 256  & 0.1   & $5 \times 10^{-4}$ & 100  & 192  & $3\times10^{-4}$ & 0.1 \\
        Darcy Flow & 200  & 4    & 0.001 & $5 \times 10^{-4}$ & 1000 & 5    & $2\times10^{-4} \rightarrow 4\times10^{-4}$ & $1 \times 10^{-4}$ \\
        PSICOV     & 200  & 8    & 0.001 & $5 \times 10^{-4}$ & 200  & 4    & $3\times10^{-4}$ & $5 \times 10^{-5}$ \\
        Cosmic     & 200  & 8    & 0.001 & $5 \times 10^{-4}$ & 150  & 16   & $0.002 \rightarrow 0.004$   & $1 \times 10^{-4}$ \\
        ECG        & 200  & 256  & 0.1   & $5 \times 10^{-4}$ & 100  & 64   & $0.001 \rightarrow 2.5\times10^{-4}$ & $1 \times 10^{-4}$ \\
        Satellite  & 200  & 4096 & 0.1   & $5 \times 10^{-4}$ & 200  & 1024 & $0.001 \rightarrow 0.002$   & $1 \times 10^{-4}$ \\
        Deepsea    & 200  & 256  & 0.1   & $5 \times 10^{-4}$ & 120  & 128  & $3\times10^{-4} \rightarrow 1.2\times10^{-3}$ & 0.01 \\
        \bottomrule
    \end{tabular}%
    }
\end{table}

\section{Search Algorithm Ablation}
\label{app:search_algorithm}

We compare three search algorithms within the same slotted search space: random search (RS) following the protocol of \citet{li2020random}, GDAS~\citep{dong2019searching}, and regularized evolution (RE)~\citep{real2019regularized} as used by AgentNAS Phase~3. All methods use the same seed architecture, training recipe, and evaluation protocol; the only difference is the search strategy. For GDAS, the virtual learning rate slot is excluded because GDAS operates over a continuous relaxation that is incompatible with this slot type. Table~\ref{tab:agentnas_search_method} reports the best non-seed architecture found by each method.

\begin{table}[h]
\centering
\caption{Search-method comparison within the AgentNAS slotted search space (lower is better). AgentNAS-RS: random search~\citep{li2020random}. AgentNAS-GDAS: GDAS~\citep{dong2019searching}, excluding the virtual learning rate slot. AgentNAS-RE: regularized evolution~\citep{real2019regularized} (AgentNAS Phase~3 default). All results report the best non-seed architecture. \textbf{Best} and \underline{second best} per row.}
\label{tab:agentnas_search_method}
\begin{tabular}{l|c|rrr}
\toprule
Task & Metric & \makecell{AgentNAS\\-RS} & \makecell{AgentNAS\\-GDAS} & \makecell{AgentNAS\\-RE} \\
\midrule
Spherical   & 0-1 err.\ (\%) & \underline{43.96} & 44.96  & \textbf{38.71} \\
NinaPro     & 0-1 err.\ (\%) & \textbf{7.28}     & \underline{7.59}  & 8.04 \\
Darcy Flow  & Rel.\ L2       & \textbf{0.0056}   & 0.0341 & \underline{0.0061} \\
Language    & 0-1 err.\ (\%) & \textbf{0.46}     & \textbf{0.46}  & \underline{0.49} \\
CIFARTile   & 0-1 err.\ (\%) & 11.28         & 75.57  & \textbf{8.48} \\
Chesseract  & 0-1 err.\ (\%) & \underline{33.51} & 34.22  & \textbf{31.39} \\
\bottomrule
\end{tabular}
\end{table}

RE and RS perform comparably overall, with neither method consistently dominating the other. RE achieves the best result on Spherical, CIFARTile, and Chesseract---tasks where the search space is large enough for evolutionary recombination to discover beneficial slot combinations. RS wins on NinaPro, Darcy~Flow, and Language, suggesting that on tasks where the performance landscape is relatively smooth or the effective search space is small, uniform sampling is sufficient. GDAS performs substantially worse on two tasks: Darcy~Flow ($0.0341$ vs.\ $0.0056$ for RS) and CIFARTile ($75.57$ vs.\ $8.48$ for RE). We have not optimized GDAS hyperparameters for the slotted setting; with tuning, the gap may narrow.

Regularized evolution and random search perform comparably, each winning on three tasks, so the gains in the main text do not depend on the choice of regularized evolution. That even random search finds competitive architectures suggests that the slotted space contains a high density of strong candidates.

\section{Other LLM Families}
\label{app:llm_families}
 
To test whether the results depend on the Claude model family, we run the pipeline with GPT-5.4 (OpenAI) and the open-weight Qwen3.6-35B-A3B on the six ablation tasks. Each model is used for all agents. Qwen3.6 is served with 8-bit quantization through OpenRouter at temperature $0.7$; GPT-5.4 does not expose a temperature setting and runs at its default. Both use a fixed split of 160 Phase~1 and 160 Phase~3 evaluations, rather than the saturation protocol of Figure~\ref{fig:ablation}. Table~\ref{tab:llm_families} shows the results next to Sonnet~4.6.
 
\begin{table}[h]
\centering
\caption{AgentNAS with different LLM families on the six ablation tasks (test metric, lower is better). LLM: AgentNAS-LLM after 160 evaluations; Full: AgentNAS-Full after 160 further NAS evaluations. Errors in \%, Darcy Flow in relative L2.}
\label{tab:llm_families}
\begin{tabular}{l|rr|rr|rr}
\toprule
& \multicolumn{2}{c|}{Sonnet 4.6} & \multicolumn{2}{c|}{GPT-5.4} & \multicolumn{2}{c}{Qwen3.6-35B-A3B} \\
Task & LLM & Full & LLM & Full & LLM & Full \\
\midrule
Spherical  & 45.08 & 38.71 & 47.03 & 44.79 & 42.24 & 42.63 \\
NinaPro    & 8.35  & 8.04  & 7.74  & 7.74  & 8.80  & 10.17 \\
Darcy Flow & 0.005 & 0.006 & 0.004 & 0.006 & 0.009 & 0.010 \\
Language   & 0.85  & 0.49  & 2.20  & 2.44  & 2.44  & 52.88 \\
CIFARTile  & 11.20 & 8.48  & 11.77 & 9.56  & 11.11 & 23.34 \\
Chesseract & 33.84 & 31.39 & 37.54 & 36.08 & 38.01 & 33.85 \\
\bottomrule
\end{tabular}
\end{table}
 
All three models produce competitive seeds in Phase~1, and Qwen3.6 finds the best seed of the three on Spherical and CIFARTile despite being much smaller. NAS improves the GPT-5.4 seeds on the same three tasks where its gain is significant for Sonnet (Spherical, CIFARTile, Chesseract), leaves NinaPro unchanged, and is worse on Darcy Flow and Language. With Qwen3.6, NAS improves only Chesseract. On Language and CIFARTile, the Slotted Architecture Planner of Qwen3.6 produced defective slot alternatives, and the resulting architectures are far worse than the seed. Together with the Haiku failure on Darcy Flow (Figure~\ref{fig:ablation}), this indicates that seed design works across a wide range of models, while search space construction needs a minimum level of capability.
\section{Full Baseline Results}
\label{app:baselines}

\def\customsuperscriptsize{0.6}
\begin{table}[h]
	\centering
	\caption{Full per-method test errors (\%, lower is better) on Unseen NAS benchmarks~\citep{geada2024insights}. Results are transcribed from the original paper~\citep{geada2024insights} and einspace~\citep{ericsson2024einspace}. \textbf{Best} and \underline{second} best performance per dataset.}
	\label{tab:unseen_results_full}
	\resizebox{\textwidth}{!}{%
		\begin{tabular}{l*{10}{@{\hspace{3pt}}c}l*{3}{@{\hspace{3pt}}c}@{\hspace{3pt}}c@{\hspace{3pt}}c}
			\toprule
			 & \multicolumn{4}{c}{Baselines} & \multicolumn{4}{c}{\makecell{Regularised evolution (RE)\\ \hspace{-2.7em} hNB201 \quad\quad\quad \texttt{einspace}}} & \multicolumn{3}{c}{Rand. Search} & \makecell{Rand.\\Sampl.} & \multicolumn{2}{c}{AgentNAS} \\
			  \cmidrule(lr){2-5} \cmidrule(lr){6-6} \cmidrule(lr){7-9} \cmidrule(lr){10-12} \cmidrule(lr){13-13} \cmidrule(lr){14-15}
			  Dataset & RN18 & \makecell{PC-\\$\text{DARTS}$} & \makecell{Dr\\$\text{NAS}$} & \makecell{Bonsai-\\$\text{Net}$} & \makecell{RE\\(Scratch)} & \makecell{RE\\(RN18)} & \makecell{RE\\(Mix)} & \makecell{RE\\(Scratch)} & $\text{DARTS}$ & $\text{Bonsai}$ & \makecell{\texttt{ein}\\\texttt{space}} & \makecell{\texttt{ein}\\\texttt{space}} & LLM & Full \\
			  \midrule
			  AddNIST     & 6.64  & 3.40  & 2.94  & \textbf{2.09}  & 6.18  & 2.46  & \underline{2.28}  & 16.13 & 2.93  & 65.83 & 33.00 & 89.87 & 2.82  & 2.42  \\
			  Language    & 7.84  & 9.88  & 11.45 & 12.35 & 7.57  & 3.16  & 2.08  & 11.88 & 9.88  & 23.17 & 12.99 & 64.74 & \underline{0.85}  & \textbf{0.49}  \\
			  MultNIST    & 8.64  & 3.32  & \textbf{1.90}  & \underline{2.83}  & 6.56  & 3.63  & 7.75  & 6.28  & 3.45  & 60.24 & 33.91 & 81.13 & 3.54  & 3.53  \\
			  CIFARTile   & 52.87 & \textbf{7.72}  & 18.92 & 8.53  & 41.69 & 39.35 & 37.24 & 69.11 & 9.26  & 75.24 & 69.10 & 74.75 & 11.20 & \underline{8.48}  \\
			  Gutenberg   & 56.68 & 50.88 & 53.38 & 51.43 & 56.30 & 45.98 & 49.84 & 63.30 & 52.28 & 71.00 & 60.42 & 80.31 & \underline{44.63} & \textbf{44.17} \\
			  GeoClassing & 9.92  & 5.39  & 3.97  & 4.34  & 7.67  & 4.69  & 4.87  & 39.57 & 4.46  & 36.44 & 30.87 & 75.65 & \underline{1.60}  & \textbf{0.97}  \\
			  Chesseract  & 40.65 & 42.80 & 41.76 & 39.24 & 36.08 & 39.69 & 38.14 & 40.50 & 40.84 & \textbf{31.17} & 38.54 & 55.17 & 33.84 & \underline{31.39} \\
			  \bottomrule
		\end{tabular}
		}
\end{table}

\subsection{Comparison with LLM-Based NAS Methods}
\label{app:llm_nas_comparison}
 
Most LLM-based NAS methods use the LLM inside the search loop over a fixed search space and are evaluated on CIFAR-10/100 or ImageNet only. Table~\ref{tab:llm_nas_comparison} places AgentNAS next to three of them on CIFAR-100, using the numbers reported in their papers. The comparison is not controlled: the methods differ in the generative model, the number of evaluated architectures, and the training protocol. We therefore offer the table as rough positioning only.
 
\begin{table}[h]
\centering
\caption{CIFAR-100 test accuracy of LLM-based NAS methods, as reported in the original papers.}
\label{tab:llm_nas_comparison}
\begin{tabular}{lrr}
\toprule
Method & Evaluated architectures & Accuracy (\%) \\
\midrule
GPT-NAS~\citep{yu2023gpt}       & 600   & 82.81 \\
LLMatic~\citep{nasir2024llmatic} & 2{,}000 & 74.17 \\
NADER~\citep{yang2025nader}     & 500   & 76.00 \\
AgentNAS                        & 320   & 84.50 \\
\bottomrule
\end{tabular}
\end{table}

\section{Architecture Families}
\label{app:families}
 
Although both benchmark suites favor convolutional networks, AgentNAS is not restricted to any family of architectures. Table~\ref{tab:families} classifies the 41 architectures produced in our experiments: the 17 main runs with Sonnet~4.6, the six ablation tasks with Haiku~4.5 and Opus~4.6 (Figure~\ref{fig:ablation}, 160-evaluation seeds), and the six ablation tasks with seeds 43 and 44. About 44\% of the Phase~1 architectures are plain or residual CNNs; the rest include Transformers, spectral neural operators, recurrent networks, and MLPs. NAS moves several seeds from convolutional to hybrid convolution--attention architectures (2 to 8).
 
\begin{table}[h]
\centering
\caption{Architecture families of the 41 seeds (P1) and final architectures (P3). Channel gating such as SE or ECA does not count as attention.}
\label{tab:families}
\resizebox{\linewidth}{!}{%
\begin{tabular}{lrrl}
\toprule
Family & P1 & P3 & Definition \\
\midrule
Residual CNN              & 14 & 13 & Convolutional network with residual connections (e.g., ResNet, WRN, PyramidNet) \\
Hybrid CNN--attention     & 2  & 8  & Convolution combined with dot-product attention \\
Transformer               & 4  & 3  & Attention only, no convolution \\
Modern CNN                & 4  & 3  & Convolution variants as the repeated unit (e.g., EfficientNet, RegNet, MobileNet) \\
Plain CNN                 & 4  & 3  & Sequential convolution and pooling without residual paths \\
TCN                       & 3  & 3  & 1D convolutions with increasing dilation \\
Spectral neural operator  & 3  & 3  & Main mixing in the Fourier domain \\
MLP / DeepSets            & 2  & 2  & No convolution or attention \\
Encoder--decoder CNN      & 3  & 1  & Multi-resolution convolutional encoder--decoder (e.g., U-Net) \\
Recurrent                 & 2  & 2  & LSTM or GRU cells \\
\midrule
Total                     & 41 & 41 & \\
\bottomrule
\end{tabular}}
\end{table}
 
\section{Benchmark Details}
\label{app:benchmarks}

\subsection{NAS-Bench-360 Dataset Descriptions}
\label{app:nasbench360}

NAS-Bench-360~\citep{tu2022bench} is a benchmark suite designed to evaluate neural architecture search methods across diverse domains. It comprises ten tasks spanning image classification, signal processing, scientific computing, and sequence analysis. Below we briefly describe each task.

\paragraph{CIFAR-100.}
A standard image classification task consisting of natural RGB images categorized into 100 fine-grained classes~\citep{krizhevsky2009learning}.

\paragraph{Spherical.}
CIFAR-100 images are projected onto the northern hemisphere of a sphere with random rotations applied, following~\citet{cohen2018spherical}. The resulting $60 \times 60$ RGB images simulate distorted signals common in omnidirectional vision and meteorological sensing.

\paragraph{NinaPro.}
A hand gesture classification task using electromyography (EMG) signals from the NinaPro DB5 dataset~\citep{atzori2012building}. EMG recordings from two Myo armbands worn by 10 subjects are sampled into 2D representations following~\citet{cote2019deep}, with 18 gesture classes.

\paragraph{FSD50K.}
A multi-label sound event classification task derived from~\citet{fonseca2021fsd50k}, containing approximately 51,000 audio clips across 200 classes from the AudioSet ontology~\citep{gemmeke2017audio}. Performance is measured by mean average precision (mAP).

\paragraph{Darcy Flow.}
A regression task that learns the mapping from initial conditions of a partial differential equation to its solution, using the Darcy Flow dataset from~\citet{li2020fourier}. Inputs and outputs are 2D grids representing fluid states at different timesteps. The evaluation metric is the relative $\ell_2$ error.

\paragraph{PSICOV.}
An inter-residue protein distance prediction task based on~\citet{adhikari2020fully}. Large-scale 2D features extracted from protein sequences serve as inputs, and the targets are pairwise distance matrices. Performance is evaluated using mean absolute error on distances below 8\,\text{\normalfont\AA} (MAE$_8$).

\paragraph{Cosmic.}
A pixel-level binary segmentation task for identifying cosmic ray artifacts in Hubble Space Telescope images of resolved galaxies~\citep{zhang2020deepcr}. The evaluation metric is $1-$AUROC.

\paragraph{ECG.}
A cardiac rhythm classification task based on the 2017 PhysioNet Challenge~\citep{clifford2017af}. ECG recordings sampled at 300\,Hz are segmented using a 1,000\,ms sliding window with 500\,ms stride, and classified into four categories (normal, disease, other, noisy). Performance is reported as F1-score.

\paragraph{Satellite.}
A satellite image time series classification task using Formosat-2 imagery over Toulouse, France~\citep{petitjean2012satellite}. Each pixel's temporal profile across multiple spectral channels is classified into 24 land cover types.

\paragraph{DeepSEA.}
A genomic sequence classification task for predicting chromatin feature activity from DNA sequences~\citep{zhou2015predicting}, based on ENCODE profiles~\citep{feingold2004encode}. Following~\citet{zhang2021ambient}, 36 out of 919 categories are used with 5\% of the training data. Performance is measured by the area under the ROC curve (AUROC).

\subsection{Unseen NAS Dataset Descriptions}
\label{app:unseennas}

Unseen NAS~\citep{geada2024insights} is a benchmark suite comprising eight tasks that test neural architecture search methods on diverse and unconventional data modalities. Tasks are categorized as Type-1 (solvable by humans) or Type-2 (requiring non-trivial pattern recognition beyond human intuition). Below we briefly describe each task.

\paragraph{AddNIST.}
A Type-1 task built from MNIST~\citep{yann_lecun_mnist_2005}. Each $3\times28\times28$ image encodes one MNIST digit per color channel, and the label is computed as $l = (r + g + b) - 1$, yielding 20 classes. The task tests whether an architecture can learn to identify digits and perform arithmetic jointly~\citep{towers_addnist_2023}.

\paragraph{Language.}
A Type-2 task where six-letter words from 10 Latin-alphabet languages are encoded as binary images via character-position mappings~\citep{towers_language_2023}. The goal is to classify each image into its source language, testing whether encoded character patterns retain sufficient linguistic information for classification.

\paragraph{MultNIST.}
A Type-1 task similar to AddNIST, with $3\times28\times28$ images whose label is $l = (r \times g \times b) \bmod 10$, producing 10 classes~\citep{towers_multnist_2023}. The modular arithmetic removes the bias toward larger digit values present in AddNIST.

\paragraph{CIFARTile.}
A Type-1 task composing four CIFAR-10~\citep{krizhevsky2009learning} images into a $3\times64\times64$ grid~\citep{towers_cifartile_2023}. The label indicates the number of distinct CIFAR-10 classes in the grid minus one, resulting in 4 classes. The task requires simultaneous recognition and comparison of multiple objects.

\paragraph{Gutenberg.}
A Type-2 task derived from Project Gutenberg literary works by six authors (Thomas Aquinas, Confucius, Hawthorne, Plato, Shakespeare, and Tolstoy)~\citep{towers_gutenberg_2023}. Consecutive three-word sequences are encoded into $1\times27\times18$ binary images via character-position mappings. The task is to predict the authorship from these spatial letter patterns.

\paragraph{Isabella.}
A Type-2 task using musical recordings from the Isabella Stewart Gardner Museum. Five-second audio snippets are converted into $1\times64\times128$ spectrograms, and the goal is to classify each into one of four composition eras (Baroque, Classical, Romantic, and 20th Century).

\paragraph{GeoClassing.}
A Type-2 task that repurposes satellite images from BigEarthNet~\citep{sumbul2019bigearthnet} by replacing ground-cover labels with the country of origin, determined by cross-referencing ESA Sentinel patch coordinates~\citep{towers_geoclassing_2023}. Each $3\times60\times60$ image is classified into one of 10 European countries based on topographic cues.

\paragraph{Chesseract.}
A Type-1 task encoding chess endgame positions from eight grandmasters as $12\times8\times8$ images via one-hot piece-type encoding~\citep{towers_chesseract_2023}. The three classes correspond to game outcomes (White wins, Draw, Black wins). The 12-channel representation tests whether architectures can handle non-standard input dimensions.

\paragraph{Blindness implementation.}
The blind setting, which follows the official protocol of the benchmark, is implemented by extending the explorer-side filter and by replacing the task identity in every prompt. Specifically, the Phase~1 task README is replaced with a synthetic minimal spec (input shape, number of classes / output shape, loss name, metric direction) under the placeholder name \texttt{TASK\_X}, and the same \texttt{TASK\_X} label is propagated to the Code Generator and Slotted Architecture Planner system prompts. The benchmark loader and the labels themselves are unchanged---blind mode only affects what the LLM agents can read, not what the harness trains on.

\paragraph{What does the LLM infer?}
\label{app:blind_recognition}
Despite the blinding, the Data Explorer often forms hypotheses about the task domain from data statistics alone. Table~\ref{tab:blind_recognition} summarizes these inferences. The LLM correctly identifies the true domain only for \textbf{Chesseract}, where it maps the 12-channel input to 2~players~$\times$~6~piece types and recognizes chess outcome prediction. For \textbf{Language}, it recovers the sequence-of-tokens abstraction without identifying the task as language identification. On the remaining five tasks, the LLM defaults to generic image classification hypotheses---treating AddNIST and MultNIST as MNIST-family variants without recognizing the multi-digit arithmetic structure, CIFARTile as standard 4-class natural-image classification, GeoClassing as a CIFAR-10/STL-10 variant, and Gutenberg as sensor or game-state data rather than character-level text. Notably, these misidentifications do not prevent the pipeline from achieving strong performance: AgentNAS reaches state-of-the-art results on Language and GeoClassing despite the LLM's incorrect or incomplete domain recognition, suggesting that the pipeline's effectiveness relies more on data-driven architecture refinement than on correct domain identification.

\begin{table}[h]
\centering
\caption{LLM domain hypotheses under the blind setting. The Data Explorer infers a task description from data statistics alone; the verdict indicates whether the true domain was identified.}
\label{tab:blind_recognition}
\resizebox{\linewidth}{!}{%
\begin{tabular}{llll}
\toprule
Task & True Domain & LLM Hypothesis & Verdict \\
\midrule
AddNIST     & 3-digit sum prediction        & MNIST-family RGB image classification       & Partial (MNIST family, missed arithmetic) \\
Chesseract  & Chess outcome (W/D/L)         & Chess outcome; 12ch = 2 players $\times$ 6 pieces & \textbf{Correct} \\
CIFARTile   & 2$\times$2 tiled CIFAR-10    & 4-class natural-image classification        & Incorrect (missed tile structure) \\
GeoClassing & Satellite geo-classification  & CIFAR-10/STL-10 variant                     & Incorrect (missed satellite imagery) \\
Gutenberg   & Authorship attribution    & Pose/sensor/game-state data                 & Incorrect (missed text domain) \\
Language    & Written language ID           & Length-24 token sequences, 1-D models       & \textbf{Almost Correct} (abstraction recovered) \\
MultNIST    & 3-digit number classification & RGB MNIST variant                           & Partial (MNIST family, missed multi-digit) \\
\bottomrule
\end{tabular}%
}
\end{table}

\paragraph{Does domain knowledge help?}
To test whether the blind setting harms performance, we run the full pipeline without blinding on four tasks and compare against the blind results (Table~\ref{tab:blind_vs_unblind}).

\begin{table}[h]
\centering
\caption{Blind vs.\ non-blind test error (\%, lower is better). Negative $\Delta$ indicates that removing blindness improves performance.}
\label{tab:blind_vs_unblind}
\begin{tabular}{lrrr}
\toprule
Task & Blind & Non-blind & $\Delta$ \\
\midrule
AddNIST    & 2.42  & \textbf{0.88}  & $-1.54$ \\
MultNIST   & 3.53  & \textbf{1.03}  & $-2.50$ \\
Gutenberg  & 44.17 & \textbf{43.93} & $-0.24$ \\
Chesseract & \textbf{31.39} & 35.34 & $+3.95$ \\
\bottomrule
\end{tabular}
\end{table}

The results are mixed. On AddNIST and MultNIST, removing blindness yields substantial gains, indicating that the LLM can leverage domain knowledge (multi-digit arithmetic) to design more effective architectures. On Gutenberg, the difference is negligible---knowing that the task is authorship classification does not translate into actionable architectural insights. Most surprisingly, on Chesseract, the blind setting \emph{outperforms} the non-blind setting by nearly 4~percentage points. We inspected the LLM logs and found that under blindness, the Data Explorer systematically benchmarks diverse proxy models (MLP, CNN, Transformer, Random Forest) before recommending an architecture, producing a richer exploration report. When the domain is known, the LLM skips this exploration and directly prescribes architectures based on domain priors, resulting in less thorough data-driven adaptation. A similar tendency appears on the other tasks where domain knowledge helps---the LLM explores less when it believes it already knows the answer. A deeper investigation of this exploration bias is interesting but is beyond the scope of this work.
\section{Implementation Details}
\label{app:impl}

This appendix documents the concrete hyperparameters and engineering choices behind the
AgentNAS pipeline. Where Appendix~\ref{app:context} describes \emph{what} each LLM agent
sees in its prompt, this appendix describes \emph{how} the surrounding pipeline is
configured: the slot
generation pipeline (\S\ref{app:impl-slot}) and the search-process protocol (\S\ref{app:impl-search}).

\subsection{Slot Generation}
\label{app:impl-slot}
\paragraph{Slot caps.}
The Slotted Architecture Planner is asked to produce 10--20 slots with 4--8 alternatives each (Appendix~\ref{app:ctx-grammar}). After parsing, the pipeline keeps at most $\texttt{max\_slots}=20$ slots in declaration order and truncates longer alternative lists to $\texttt{max\_alternatives\_per\_slot}=8$.

\subsection{Search Process}
\label{app:impl-search}

\paragraph{Phase~1 budget and patience.}
Phase~1 has a hard cap of $\texttt{max\_evaluations}=160$ architecture evaluations and a
patience counter of $\texttt{patience}=64$ \emph{successful} consecutive
no-improvement evaluations. A failed evaluation
(\texttt{rejected}/\texttt{oom}/\texttt{error}/\texttt{timeout}) consumes one unit of the
$160$-evaluation budget but does \emph{not} increment patience. The two limits serve different purposes: the budget protects against runaway compute, while patience protects against
runaway wall-clock with no metric movement; counting failures toward patience would let a
streak of OOM proposals trigger early termination on a task where the LLM simply
needed a few VRAM-aware retries. After every batch the planner state and per-evaluation
log are checkpointed to \texttt{state.json} so that a pod restart resumes exactly where
the previous process left off.

\paragraph{Phase~3 budget and patience.}
Phase~3 reuses the same $\texttt{max\_evaluations}=160$ pool and has its own independent
patience counter $\texttt{nas\_patience}=64$, also counted only over successful
evaluations. Evolution additionally caps generations at $\texttt{max\_generations}=10$
(initial population $+$ $9$ offspring waves). With $\texttt{population\_size}=16$ this
gives a soft upper bound of $160$ total evaluations, which lines up with the budget
ceiling; in practice the patience trigger fires first on tasks where evolution converges
within a few generations.

\paragraph{Proxy training scaling.}
The Planner is told to design its training recipe for the \emph{full} schedule. The
pipeline then runs the proxy at $\texttt{proxy\_epoch\_fraction}=0.3$, i.e. proxy
\texttt{epochs}~$=\max(1, \mathrm{round}(0.3 \cdot E_{\mathrm{full}}))$. The
\texttt{warmup\_epochs} and any time-based scheduler parameters (\texttt{T\_max} for cosine,
\texttt{step\_size} for step LR, \texttt{patience} for plateau) are scaled by the same
factor and floored at $1$, so the LR schedule still completes within the shorter run.
On top of the epoch cap, every proxy run is subject to a $60$-minute GPU time hard
cap (\texttt{proxy\_time\_limit\_s}~$=3600$\,s), measured via paired CUDA events around
each forward/backward step rather than wall clock; data-loading time and evaluation time
are excluded so that a slow disk does not penalize an architecture's measured fitness.
The same VRAM cap (\texttt{max\_vram\_mb}~$=11{,}264$, the RTX~2080~Ti's $11$\,GB) is
enforced via \texttt{torch.cuda.max\_memory\_allocated} sampled per epoch; an OOM exception
during training is caught and reported as \texttt{status="oom"}.

\paragraph{Final training: two waves.}
After Phase~1 (and again after Phase~3), the top-$8$ architectures by proxy validation
metric are passed through a two-wave full-training protocol. We use $\texttt{top\_k\_final}=8$
to fully utilize the $8$~GPUs of one node in parallel.

\begin{itemize}
  \item \textbf{Wave~1: validation metric.} Each architecture is retrained with the
        \emph{full} epoch schedule (\texttt{proxy\_epoch\_fraction}~$=1.0$) on the
        train split, using the validation split for best-checkpoint selection. The harness
        records, for every epoch, the validation metric and the model's state dict; only
        the state dict at the best validation epoch is kept. The Wave~1 metric is what every downstream
        pipeline decision uses.
  \item \textbf{Wave~2: test metric.} The same architectures are then retrained
        from scratch on the train$+$validation \emph{combined} set (no held-out split for
        checkpoint selection), using the \emph{last} checkpoint, and evaluated once on
        the held-out test set. This wave is run only for analysis and is the test column
        in the main paper.
\end{itemize}

Both waves use the same $\texttt{final\_time\_limit\_s}=86{,}400$\,s ($24$\,h) GPU-time
hard cap, again measured via CUDA events. Failed Wave~1 architectures are reported as
\texttt{status="error"} or \texttt{"oom"}/\texttt{"timeout"}; the Wave~1 winner is chosen
as the lowest \texttt{val\_final\_metric} among successful entries.

\paragraph{Population initialization for evolution.}
The Phase~3 evolutionary search initializes its population of $16$ as: the seed
($1$~individual, choices identical to \texttt{seed\_choices}), then $\lfloor (16-1)/3\rfloor=5$
single-slot mutations of the seed, $5$ double-slot mutations of the seed, and the
remaining $5$ uniformly-random configurations. For the \emph{random-search} ablation, the population is instead drawn as $16$ i.i.d.\
uniform samples from the search space.

\paragraph{Regularized Evolution.}
We perform regularized evolution through tournament selection with $\texttt{tournament\_size}=5$ followed by $50/50$ single/double-slot mutation. After evaluation, the population is updated by appending offspring and removing the oldest
individuals until size $16$ is restored; ages are incremented at end-of-generation.

\subsection{Modularization Ablation}
\label{app:modularization}

\begin{table}[h]
\centering
\caption{Effect of modularization on tasks with block-stack seeds (lower is better). Without modularization, NAS fails to improve upon the seed: validation metric never improves and patience is exhausted.}
\label{tab:modularization}
\begin{tabular}{l|c|cc}
\toprule
Task & Phase 1 & \makecell{Without\\modularization} & \makecell{With\\modularization} \\
\midrule
CIFAR-100 & 16.74 & 18.15 & \textbf{15.50} \\
CIFARTile & 11.20 & 14.78 & \textbf{8.48} \\
\bottomrule
\end{tabular}
\end{table}

As described in Section~\ref{sec:phase2}, the Slot Planner applies a modularization step when the seed uses a block-stack pattern, grouping tightly coupled primitives into coherent modules before decomposition. This step is applied selectively and automatically; it does not alter the seed's architecture but prevents Phase~2 from fragmenting coupled operation sequences, which can degrade the quality of the resulting search space. Table~\ref{tab:modularization} describes two cases where modularization is activated whose seeds exhibit block-stack structure. Without modularization, the search space degenerates: NAS exhausts its patience criterion early and produces weaker architectures. With modularization, NAS utilizes the full budget and delivers substantial improvements on both tasks.

\section{Context Management and Prompts for Agents}
\label{app:context}

This appendix details, for each LLM agent in the AgentNAS pipeline, what enters its context window at every call. Five agents are involved across the three phases:

\begin{itemize}
  \item \textbf{Planner} (\S\ref{app:ctx-planner}) — Phase~1 controller; stateful, multi-turn.
  \item \textbf{Data Explorer} (\S\ref{app:ctx-explorer}) — Phase~1 sandbox; autoregressive code-execution loop.
  \item \textbf{Code Generator} (\S\ref{app:ctx-executor}) — Phase~1 and Phase~2; translates descriptions into PyTorch.
  \item \textbf{Modularizer} (\S\ref{app:ctx-modularizer}) — Phase~2; refactors the seed into per-stage \texttt{nn.Module}s.
  \item \textbf{Slotted Architecture Planner} (\S\ref{app:ctx-grammar}) — Phase~2; single-call slotted-architecture designer.
\end{itemize}

\subsection{Planner}
\label{app:ctx-planner}

\begin{tcolorbox}[breakable, enhanced, colback=gray!5, colframe=black!55, fonttitle=\bfseries, title={Planner system prompt}]
\begin{Verbatim}[fontsize=\footnotesize, breaklines=true, breakanywhere=true]
You are the Planner agent for a neural architecture search pipeline. Your goal
is to design the best possible neural network architecture for the given task.

## Available Actions
At each turn, choose ONE action:

1. <EXPLORE> {purpose} </EXPLORE>
   Launch the Data Explorer to analyze the dataset.

2. <EXPERIMENT>
   Propose up to {n_gpu} architectures for parallel evaluation. For each:
     <ARCH id="arch_NNN">
       <DESCRIPTION>      {high-level architecture description}    </DESCRIPTION>
       <REASONING>        {why this design, what you expect}       </REASONING>
       <TRAINING_RECIPE>  {optimizer, lr, scheduler, epochs, ...}  </TRAINING_RECIPE>
     </ARCH>
   </EXPERIMENT>
   To modify a prior architecture, name it by literal arch_NNN id in
   DESCRIPTION; the pipeline regex-injects its full code into the Executor.

3. <STOP>
   <FINAL_SELECTION> {arch_ids for full training} </FINAL_SELECTION>
   </STOP>

After receiving experiment results:
  <SUMMARY> {what worked, what failed, surprises, next direction} </SUMMARY>

## Constraints
- {vram_mb} MB VRAM per GPU
- Proxy training: {proxy_epoch_pct}% of proposed epochs, {time_limit_min}-min GPU cap
- Final training: full proposed epochs, 24-hour GPU cap
- {budget_remaining} of {budget_total} architecture evaluations remaining
- Patience: {patience_counter}/{patience} consecutive no-improvement
- NO pre-trained weights. All models trained from scratch.

## Performance Baseline
{sota_info}    Lower metric is better.

## Training-recipe whitelist (anything else is silently dropped)
- optimizer:    sgd | adam | adamw                      (others -> AdamW)
- scheduler:    cosine | step | warmup_cosine | plateau | none
- augmentations (harness-level): subset of {mixup, cutmix, random_erasing}
- loss: TASK-FIXED. Only label_smoothing and mixup/cutmix are loss-side knobs.
- NOT supported: AMP / fp16 / bf16, torch.compile, grad accum/checkpointing.
- For task-specific augs (e.g. time-shift), implement INLINE in
  Architecture.forward() gated on self.training.

## Task
{task_readme}

## Data Explorer Report
{explorer_report}

\end{Verbatim}
\end{tcolorbox}

\subsection{Data Explorer}
\label{app:ctx-explorer}

\begin{tcolorbox}[breakable, enhanced, colback=gray!5, colframe=black!55, fonttitle=\bfseries, title={Data Explorer system prompt}]
\begin{Verbatim}[fontsize=\footnotesize, breaklines=true, breakanywhere=true]
You are a Data Explorer agent for a neural architecture search pipeline. Your
job is to analyze a dataset by writing and executing Python code.

## Environment
- Available: numpy, pandas, scipy, matplotlib, seaborn, librosa, h5py,
  sklearn, torch, and the standard library.
- Data is at: {data_root}/{data_subdir}/
- You are working in a temporary directory. Save plots to ./plot.png.

## Workflow
At each step, write Python code in one or more ```python``` blocks. ALL blocks
in a single turn are concatenated and executed together in one fresh
subprocess (notebook-style WITHIN a turn; variables/imports do NOT carry
across turns -- each turn starts a new process). stdout+stderr (truncated to
5000 chars) returns as the next user message.

CRITICAL rules:
- NEVER fabricate "**Execution output:**" or any output block in your own
  message -- outputs come back to you in the NEXT user turn.
- Do NOT produce <REPORT> until you have actually executed at least one
  ```python``` block and observed real output.

When done, write:
  <REPORT> {dataset stats; key characteristics; domain observations;
            recommendations for the Planner} </REPORT>

## Constraints
- Maximum {max_steps} steps. On step {max_steps}, <REPORT> is mandatory.
- Keep code concise and focused on the investigation purpose.

## Purpose
{purpose}      # provided by the Planner via <EXPLORE>
\end{Verbatim}
\end{tcolorbox}

\subsection{Code Generator}
\label{app:ctx-executor}

\begin{tcolorbox}[breakable, enhanced, colback=gray!5, colframe=black!55, fonttitle=\bfseries, title={Code Generator system prompt}]
\begin{Verbatim}[fontsize=\footnotesize, breaklines=true, breakanywhere=true]
You are a Code Executor for a neural architecture search pipeline. Your job
is to translate architecture descriptions into working PyTorch code.

## What you produce
1. Model code: a class `Architecture` extending `nn.Module`, with
   __init__(self) and forward(self, x).
   - x shape: (batch, {input_shape})
   - output shape: (batch, {output_spec})
   - Imports: torch, torch.nn, torch.nn.functional, stdlib only.
   - NO pre-trained weights. No torchvision.models, no timm, no checkpoints.

2. Training config: JSON dict (schema below). Use ONLY the fields listed.

Wrap model in ```python```, config in ```json```.

## Constraints
- VRAM cap: {vram_mb} MB.
- Proxy: {proxy_epoch_pct}% of proposed epochs, {time_limit}-min GPU cap.
  Final: full proposed epochs, 24-hour GPU cap.
- Loss: {loss_info}    # CrossEntropyLoss / BCEWithLogitsLoss / regression
  Output raw logits/predictions; do NOT apply softmax/sigmoid.
- On retry: only fix bugs (shape mismatches, runtime errors). Do NOT change
  the architecture's functional design.

## Training-recipe whitelist (anything else is silently dropped)
- optimizer:    sgd | adam | adamw                       (others -> AdamW)
- scheduler:    cosine | step | warmup_cosine | plateau | none
- augmentations: subset of {mixup, cutmix, random_erasing}
- NOT supported: amp/autocast, torch.compile, grad accumulation
- For unsupported augs (time_shift, gaussian_noise, ...), implement INLINE
  in forward() gated on self.training:
    if self.training:
        # e.g. shift = int(torch.randint(-4, 5, (1,)).item())
        # x = torch.roll(x, shifts=shift, dims=-1) + torch.randn_like(x)*0.01
        ...

## Training config fields honored by the harness
{
  "optimizer": "adamw",
  "lr": ..., "weight_decay": ..., "momentum": ..., "betas": ...,
  "scheduler": "cosine", "scheduler_params": {...}, "warmup_epochs": ...,
  "batch_size": ..., "epochs": ...,
  "augmentations": [...], "augmentation_params": {...},
  "grad_clip": ..., "label_smoothing": ...
}

## Task
{task_description}
\end{Verbatim}
\end{tcolorbox}

\begin{tcolorbox}[breakable, enhanced, colback=gray!5, colframe=black!55, fonttitle=\bfseries, title={Code Generator input prompt}]
\begin{Verbatim}[fontsize=\footnotesize, breaklines=true, breakanywhere=true]
## Recent Architectures (code for reference)
# only included if the Planner's description mentions arch_NNN tokens; the
# pipeline regex-matches them and injects the matching code blocks here.
The Planner's description may reference these by `arch_NNN` -- use them as
the baseline to modify.

### {arch_id} (status={...}, metric={...})
```python
{full_prior_architecture_code}
```
... (one block per referenced architecture)

## Architecture to implement
{architecture_description}    # from <DESCRIPTION> in the Planner proposal

## Training recipe
{training_recipe}             # from <TRAINING_RECIPE> in the Planner proposal

Produce the `Architecture` class and training config JSON.

# On validation failure, the truncated traceback is fed back as the next user
# turn (up to 10 retries):
The model failed validation with this error:
```
{truncated_traceback}
```

Please fix the code. Only fix bugs -- do NOT change the architecture design.
\end{Verbatim}
\end{tcolorbox}

\begin{tcolorbox}[breakable, enhanced, colback=gray!5, colframe=black!55, fonttitle=\bfseries, title={Code Generator system prompt}]
\begin{Verbatim}[fontsize=\footnotesize, breaklines=true, breakanywhere=true]
You are implementing a PyTorch nn.Module for a neural architecture search slot.

## What to produce
A complete Python class named `{class_name}` that extends `nn.Module`.

```python
class {class_name}(nn.Module):
    def __init__(self):
        super().__init__()
        # ... your layers here (NO constructor arguments; dimensions hardcoded
        #     from this slot's input/output shape) ...

    {forward_signature}:    # forward(self, x)            for "single"
                            # forward(self, x, residual)  for "dual"
        # x shape:        (batch, {input_shape_at_slot})
        # residual shape: (batch, {residual_shape})       # "dual" only
        # return shape:   (batch, {output_shape_at_slot})
        return x
```

## Interface
{interface_note}    # explanation of single vs. dual semantics for this slot

## Slot:        {slot_description}
## Alternative: {alternative_description}

## Constraints
- Imports: torch, torch.nn, torch.nn.functional only.
- NO pre-trained weights. Random initialization only.
- Class must be fully self-contained.
- Wrap code in a ```python``` block.

# User-side message (one line):
Implement `{class_name}` for: {alternative_description}

# Retry feedback on validation failure (up to 10 retries):
The module failed validation:
```
{truncated_traceback}
```
Fix it. Class: `{class_name}`, forward signature: `{forward_signature}`.
\end{Verbatim}
\end{tcolorbox}

\subsection{Modularizer}
\label{app:ctx-modularizer}

\begin{tcolorbox}[breakable, enhanced, colback=gray!5, colframe=black!55, fonttitle=\bfseries, title={Modularizer system prompt}]
\begin{Verbatim}[fontsize=\footnotesize, breaklines=true, breakanywhere=true]
You are a Modularizer agent. You refactor a PyTorch Architecture's source
code so that each repeated stage is wrapped in its own self-contained
nn.Module subclass. Forward-pass behavior must be IDENTICAL to the original
(bit-equivalent under torch.allclose(rtol=1e-5, atol=1e-6) in eval mode after
the original's state_dict is transplanted into the refactored module).

## What you must produce
1. All helper classes the original used (RegNetBlock, FNOBlock, ...).
   Copy them VERBATIM -- no renames, no reordering, no layer-construction
   changes.

2. NEW per-stage Stage classes (Stage1, Stage2, Stage3, ...). Each must:
   - Have a docstring stating the contract, e.g.:
     '''INTERNAL: BatchNorm + ReLU + SE + residual baked in. Outer
     norm/act/skip slots SHOULD be Identity.'''
   - In __init__, instantiate blocks/parameters in the SAME order with the
     SAME hyperparameters as the original stage construction.
   - In forward(self, x), perform exactly the same operations the original
     did for this stage -- INCLUDING any per-stage outer norm/activation/
     residual previously written in the top-level Architecture.forward.
   - Single tensor in / single tensor out.

3. Rewritten Architecture class:
   - Same helper attributes (stem, pool, head, ...) constructed identically.
   - Each repeated stage replaced by `self.stageN = StageN()` (instead of
     `nn.ModuleList([...])`).
   - Architecture.forward simplified to:
       # data-augmentation block (UNCHANGED, copy verbatim)
       x = self.stem(x)
       x = self.stage1(x); x = self.stage2(x); ...
       x = self.pool(x).flatten(1)
       return self.head(x)

## Hard constraints (any violation rejects the output)
- No parameter add/remove. sum(p.numel()) must match exactly.
- No op-type changes (no ReLU<->GELU, BN<->LN, Conv<->DWConv, etc.).
- No op-order changes within a stage.
- Data-augmentation block: copy verbatim.
- _init_weights logic: copy verbatim.
- Submodule construction order in __init__: must match original (RNG align).
- No new imports beyond what original used.

## Output format
Return ONLY a single ```python``` block containing the complete refactored
source. No prose outside the code block.

## Original source
```python
{original_code}
```

## Task description
{task_description}    # "Task: {task_name}, Input shape: (batch, ...)"

# Initial user-side message:
Refactor the Architecture above per the contract. Wrap each repeated stage in
its own Stage class with internal norm/act/skip baked in, simplify the
top-level forward, and ensure forward-pass equivalence. Return ONE
```python``` block.

# Retry feedback on validation failure (up to 5 retries):
Validation FAILED:

{failure_summary}
# one of, e.g.:
# - "Parameter count mismatch: original={...} vs modularized={...}. ..."
# - "Parameter shapes differ as a sorted multiset. ..."
# - "Forward output mismatch: max_abs_diff={diff}. After transplanting the
#    original weights into your refactored module, the forward pass produced
#    different outputs in eval mode. Check op order, per-stage outer
#    norm/act/skip lifted into the stage's forward, and missing operations."

Re-emit the refactored Architecture, fixing the issue. The contract (no param
adds/removes, no op-type or op-order changes, identical construction order,
augmentation block untouched) must be respected.
\end{Verbatim}
\end{tcolorbox}

\subsection{Slotted Architecture Planner}
\label{app:ctx-grammar}

\begin{tcolorbox}[breakable, enhanced, colback=gray!5, colframe=black!55, fonttitle=\bfseries, title={Slotted Architecture Planner system prompt}]
\begin{Verbatim}[fontsize=\footnotesize, breaklines=true, breakanywhere=true]
You are a Grammar Planner for neural architecture search. Your job is to
design a rich, domain-aware search space around a seed architecture.

## What you produce
A JSON object with `scaffold_code` and `slots`.

### scaffold_code
The main Architecture class with searchable components replaced by slot calls:
- self.xxx = SLOT_xxx()  in __init__   -- NO constructor arguments. Always
  SLOT_xxx() with no args. Dimensions are hardcoded inside each module from
  the slot's input/output shape.
- forward calls:
    "single" slot:  x = self.xxx(x)
    "dual"   slot:  x = self.xxx(x, residual)
- Do NOT include `class SLOT_xxx` definitions in scaffold_code. The actual
  class bodies are inserted automatically before the scaffold during
  assembly. Stub forwards = silent identity collapse.
- Every slot you instantiate in __init__ must be called from forward
  (or from a helper method that forward invokes).

### slots
Define 10-20 fine-grained slots:
- Per-stage operation slots: stageN_op, stageN_norm, stageN_act,
                              stageN_attn, stageN_pool
- Cross-cutting parameter slots: width_stageN, kernel_stageN
Skip/composition can live inside op modules OR as separate "dual" slots.

## Designing good alternatives
Each slot: 4-8 alternatives, including
1. Seed choice (always first).
2. Domain-specific novel ops -- e.g.:
   - audio:    frequency-band processing, multi-scale temporal conv
   - images:   multi-scale feature extraction, deformable conv
   - 1D sigs:  dilated causal conv, wavelet-inspired blocks
3. Standard alternatives (conv3x3, conv1x1, depthwise-separable,
   squeeze-excite, self-attention).
4. Identity (pass-through) and Zero (outputs zeros) -- let NAS discover
   that a component is unnecessary.

## Slot JSON format
single-input:
{
  "name": "stageN_op", "class_name": "SLOT_stageN_op",
  "description": ..., "input_shape": [...], "output_shape": [...],
  "interface": "single", "seed_description": ...,
  "alternatives": [...]
}
dual-input (skip/composition):
{ ..., "interface": "dual", "residual_shape": [...] }

## Key principles
- More slots = richer search. 10-20 is the target.
- Domain creativity matters; do not just list standard ConvNet blocks.
- Include identity AND zero for EVERY slot.
- Input/output shapes must be EXACT.
- NO pre-trained weights.

## Seed Architecture
```python
{seed_code}
```

## Seed Performance
Metric: {val_final_metric}, Params: {param_count}

## Data Explorer Report
{explorer_report}     # truncated to 3000 chars

## Planner's Accumulated Insights
{planner_summary}     # last 3000 chars of the summaries (recency-preserving)

## Task
{task_description}    # blind-ablation runs replace task name with TASK_X

## Recognizing self-contained stage modules
If the seed defines a stage as a self-contained nn.Module (with internal
norm/act/residual), then:
- Create ONE op slot per stage (the stage as a whole).
- Create OUTER norm/act/skip slots BUT mark their alt0 explicitly as Identity
  with the description "Identity (already inside the stage block)".
- Provide non-Identity alternatives for the outer slots so NAS can ADD an
  outer norm/act if useful, but the seed pick is Identity.
DO NOT unwrap the stage into multiple sub-slots that duplicate functionality
already inside the block.
\end{Verbatim}
\end{tcolorbox}


\end{document}